%% file: neurips_2020.tex
\documentclass{article}

\usepackage[final,nonatbib]{neurips_2020}
\usepackage{microtype}
\usepackage{graphicx}
\usepackage{booktabs}
\usepackage{hyperref}

\usepackage[utf8]{inputenc}
\usepackage[T1]{fontenc}
\usepackage{adjustbox}
\usepackage{algorithm}
\usepackage[noend]{algorithmic}
\usepackage{amsfonts}
\usepackage[font=scriptsize]{subfig}
\usepackage{amsmath}
\usepackage{amssymb}
\usepackage{balance}
\usepackage{bbm}
\usepackage{bm}
\usepackage[font=small]{caption}
\usepackage{cite}
\usepackage{clipboard}
\usepackage{enumitem}
\usepackage{float}
\usepackage{inconsolata}
\usepackage{makecell}
\usepackage{mathbbol}
\usepackage{mathtools, nccmath}
\usepackage{microtype}
\usepackage{multirow}
\usepackage{nicefrac}
\usepackage{pifont}
\usepackage{setspace}
\usepackage{thmtools}
\usepackage{thm-restate}
\usepackage{url}
\usepackage{wrapfig}
\usepackage{xcolor}

\DeclareSymbolFontAlphabet{\mathbb}{AMSb}
\DeclareSymbolFontAlphabet{\mathbbl}{bbold}

\DeclareMathOperator*{\argmax}{argmax}
\DeclareMathOperator*{\argmin}{argmin}

\definecolor{faintgray}{gray}{0.9}

\newcommand{\pix}{\kern 0.1em}
\newcommand{\pmm}{\kern 0.35em$\pm$\kern 0.35em}
\newcommand{\cm}{\ding{51}}
\newcommand{\xm}{{\color{faintgray}\ding{55}}}
\newcommand{\COM}[1]{\hfill$\triangleright$ #1}

\newcommand{\QED}{\hfill\raisebox{-0.5pt}{\scalebox{0.88}{$\square$}}}
\newcommand{\dayum}[1]{{#1\parfillskip=0pt\par}}

\declaretheorem[name=Proposition]{reproposition}
\declaretheorem[name=Lemma,numberlike=reproposition]{relemma}

\title{Strictly Batch Imitation Learning \\
\scalebox{0.9}{by} Energy-based Distribution Matching}

\author{%
~~~~~~~~~~~~~~~\pix
Daniel Jarrett\pix\raisebox{-0.5pt}{\scalebox{0.9}{$^{*}$}}
~~~~~~~~~~~~~~~~~~~~~~~~~
Ioana Bica\pix\raisebox{-0.5pt}{\scalebox{0.9}{$^{*}$}}
~~~~~~~~~~~~~~~~~
Mihaela van der Schaar \\
~~~~~~
University of Cambridge
~~~~~~~~~~~~
University of Oxford
~~~~~~~~~~~~
University of Cambridge \\
{\small
\texttt{daniel.jarrett@maths.cam.ac.uk}
}
~\pix
The Alan Turing Institute
\pix
University\pix of California,\pix Los\pix Angeles \\
{\small
~~~~~~~~~~~~~~~~~~~~~~~~~~~~~~~~~~~~~~~~~~~~~~~~~~~~~~~~~~~~~~
\texttt{ioana.bica@eng.ox.ac.uk}
~~~~~~~~}
The Alan Turing Institute \\
{\small
~~~~~~~~~~~~~~~~~~~~~~~~~~~~~~~~~~~~~~~~~~~~~~~~~~~~~~~~~~~~~~
~~~~~~~~~~~~~~~~~~~~~~~~~~~~~~~~~~~~~~~~~~~~~~~~~~~~~~~~~~~~~~
~
\texttt{mv472@cam.ac.uk}
}
}

\begin{document}

\maketitle

\vspace{-2em}\begin{abstract}

Consider learning a policy purely on the basis of demonstrated behavior\textemdash that is, with no access to reinforcement signals, no knowledge of transition dynamics, and no further interaction with the environment. This \textit{strictly batch imitation learning} problem arises wherever live experimentation is costly, such as in healthcare. One solution is simply to retrofit existing algorithms for apprenticeship learning to work in the offline setting. But such an approach leans heavily on off-policy evaluation or offline model estimation, and can be indirect and inefficient. We argue that a good solution should be able to explicitly parameterize a policy (i.e. respecting action conditionals), implicitly learn from rollout dynamics (i.e. leveraging state marginals), and\textemdash crucially\textemdash operate in an entirely offline fashion. To address this challenge, we propose a novel technique by \textit{energy-based distribution matching} (EDM): By identifying parameterizations of the (discriminative) model of a policy with the (generative) energy function for state distributions, EDM yields a simple but effective solution that equivalently minimizes a divergence between the occupancy measure for the demonstrator and a model thereof for the imitator. Through experiments with application to control and healthcare settings, we illustrate consis- tent performance gains over existing algorithms for strictly batch imitation learning.

\end{abstract}\section{Introduction}

Imitation learning deals with training an agent to mimic the actions of a demonstrator. In this paper, we are interested in the specific setting of \textit{strictly batch imitation learning}\textemdash that is, of learning a policy purely on the basis of demonstrated behavior, with no access to reinforcement signals, no knowledge of transition dynamics, and\textemdash importantly\textemdash no further interaction with the environment. This problem arises wherever live experimentation is costly, such as in medicine, healthcare, and industrial processes. While behavioral cloning is indeed an intrinsically offline solution as such, it fails to exploit precious information contained in the distribution of states visited by the demonstrator.

\dayum{Of course, given the rich body of recent work on (online) apprenticeship learning, one solution is simply to repurpose such existing algorithms\textemdash including classic inverse reinforcement learning and more recent adversarial imitation learning methods\textemdash to operate in the offline setting. However, this strategy leans heavily on off-policy evaluation (which is its own challenge per se) or offline model estimation (inadvisable beyond small or discrete models), and can be indirect and inefficient---via off-policy alternating optimizations, or by running RL in a costly inner loop. Instead, we argue that a good solution should directly parameterize a policy (i.e. respect action conditionals), account for rollout dynamics (i.e. respect state marginals), and\textemdash crucially\textemdash operate in an entirely offline fashion without recourse to off-policy evaluation for retrofitting existing (but intrinsically online) methods.}

\textbf{Contributions}~
In the sequel, we first formalize imitation learning in the \textit{strictly batch} setting, and motivate the unique desiderata expected of a good solution (Section \ref{sec:sbil}). To meet this challenge, we propose a novel technique by \textit{energy-based distribution matching} (EDM) that identifies parameterizations of the (discriminative) model of a policy with the (generative) energy function for state distributions (Section \ref{sec:edm}). To understand its relative simplicity and effectiveness for batch learning, we relate the EDM objective to existing notions of divergence minimization, multitask learning, and classical imitation learning (Section \ref{sec:discussion}). Lastly, through experiments with application to control tasks and health- care, we illustrate consistent improvement over \mbox{existing algorithms for offline imitation (Section \ref{sec:experiments})}.

\section{Strictly Batch Imitation Learning}\label{sec:sbil}

\textbf{Preliminaries}~
We work in the standard Markov decision process (MDP) setting, with states $s\in\mathcal{S}$, actions $a\in\mathcal{A}$, transitions $T\in\Delta(\mathcal{S})^{\mathcal{S}\times\mathcal{A}}$, rewards $R\in\mathbb{R}^{\mathcal{S}\times\mathcal{A}}$, and discount $\gamma$. Let $\pi\in\Delta(\mathcal{A})^{\mathcal{S}}$ denote a policy, with induced occupancy measure $\rho_{\pi}(s,a)\doteq\mathbb{E}_{\pi}[\sum_{t=0}^{\infty}\gamma^{t}\mathbbl{1}_{\{s_{t}=s,a_{t}=a\}}]$, where the expectation is understood to be taken over $a_{t}\sim\pi(\cdot|s_{t})$ and $s_{t+1}\sim T(\cdot|s_{t},a_{t})$ from some initial distribution. We shall also write $\rho_{\pi}(s)\doteq\sum_{a}\rho_{\pi}(s,a)$ to indicate the state-only occupancy measure. In this paper, we operate in the most minimal setting where neither the environment dynamics nor the reward function is known. Classically, \textit{imitation learning} \cite{le2016smooth, hussein2017imitation, yue2018imitation} seeks an imitator policy $\pi$ as follows:
\begin{equation}\label{eqn:obj_il}
\textstyle\argmin_{\pi}\mathbb{E}_{s\sim{\rho_\pi}}\mathcal{L}\big(\pi_{D}(\cdot|s),\pi(\cdot|s)\big)
\end{equation}
\dayum{where $\mathcal{L}$ is some choice of loss. In practice, instead of $\pi_{D}$ we are given access to a sampled dataset $\mathcal{D}$ of state-action pairs $s,a\sim\rho_{D}$. (While here we only assume access to \textit{pairs}, some algorithms require \textit{triples} that include next states). Behavioral cloning \cite{pomerleau1991efficient, bain1999framework, syed2010reduction} is a well-known (but naive) approach that simply ignores the endogeneity of the rollout distribution, replacing $\rho_{\pi}$ with $\rho_{D}$ in the expectation. This reduces imitation learning to a supervised classification problem (popularly, with cross-entropy loss), though the potential disadvantage of disregarding the visitation distribution is well-studied \cite{ross2010efficient, melo2010learning, piot2014boosted}.}

\textbf{Apprenticeship Learning}~
To incorporate awareness of dynamics, a family of techniques (commonly referenced under the ``apprenticeship learning'' umbrella) have been developed, including classic inverse reinforcement learning algorithms and more recent methods in adversarial imitation learning. Note that the vast majority of these approaches are \textit{online} in nature, though it is helpful for us to start with the same formalism. Consider the (maximum entropy) reinforcement learning setting, and let $R_{t}\doteq R(s_{t},a_{t})$ and $\mathcal{H}_{t}\doteq$$-$$\log\pi(\cdot|s_{t})$. The (forward) primitive $\text{RL}:\mathbb{R}^{\mathcal{S}\times\mathcal{A}}\rightarrow\Delta(\mathcal{A})^{\mathcal{S}}$ is given by:
\begin{equation}\label{eqn:obj_rl}
\text{RL}(R)\doteq\textstyle\argmax_{\pi}\Big(\mathbb{E}_{\pi}[\sum_{t=0}^{\infty}\gamma^{t}R_{t}]+H(\pi)\Big)
\end{equation}
where (as before) the expectation is understood to be taken with respect to $\pi$ and the environment dynamics, and $H(\pi)\doteq\mathbb{E}_{\pi}[\sum_{t=0}^{\infty}\gamma^{t}\mathcal{H}_{t}]$. A basic result \cite{ziebart2010modeling, haarnoja2017reinforcement} is that the (soft) Bellman operator is contractive, so its fixed point (hence the optimal policy) is unique. Now, let $\psi:\mathbb{R}^{\mathcal{S}\times\mathcal{A}}\rightarrow\mathbb{R}$ denote a reward function regularizer. Then the (inverse) primitive $\text{IRL}_{\psi}:\Delta(\mathcal{A})^{\mathcal{S}}\rightarrow\mathcal{P}(\mathbb{R}^{\mathcal{S}\times\mathcal{A}})$ is given by:
\begin{equation}\label{eqn:obj_irl}
\text{IRL}_{\psi}(\pi_{D})\doteq\textstyle\argmin_{R}\Big(\psi(R)+\textstyle\max_{\pi}\big(\mathbb{E}_{\pi}[\sum_{t=0}^{\infty}\gamma^{t}R_{t}]+H(\pi)\big)-\mathbb{E}_{\pi_{D}}[\sum_{t=0}^{\infty}\gamma^{t}R_{t}]\Big)
\end{equation}
Finally, let $\tilde{R}\in\text{IRL}_{\psi}(\pi_{D})$ and $\pi=\text{RL}(\tilde{R})$, and denote by $\psi^{*}:\mathbb{R}^{\mathcal{S}\times\mathcal{A}}\rightarrow\mathbb{R}$ the Fenchel conjugate of regularizer $\psi$. A fundamental result \cite{ho2016generative} is that ($\psi$-regularized) apprenticeship learning can be taken as the composition of forward and inverse procedures, and obtains an imitator policy $\pi$ such that the induced occupancy measure $\rho_{\pi}$ is close to $\rho_{D}$ as determined by the (convex) function $\psi^{*}$:
\begin{equation}\label{eqn:obj_rl_irl}
\text{RL}\circ\text{IRL}_{\psi}(\pi_{D})=\textstyle\argmax_{\pi}\Big(-\psi^{*}(\rho_{\pi}-\rho_{D})+H(\pi)\Big)
\end{equation}
Classically, IRL-based apprenticeship methods \cite{ng2000algorithms, abbeel2004apprenticeship, neu2007apprenticeship, ramachandran2007bayesian, syed2008game, ziebart2008maximum, babes2011apprenticeship, choi2011map, jarrett2020inverse} simply execute $\text{RL}$ repeatedly in an inner loop, with fixed regularizers $\psi$ for tractability (such as indicators for linear and convex function classes). More recently, adversarial imitation learning techniques leverage Equation \ref{eqn:obj_rl_irl} (modulo $H(\pi)$, which is generally less important in practice), instantiating $\psi^{*}$ with various \mbox{$\phi$-divergences \cite{ho2016generative, baram2017model, jeon2018bayesian, finn2016connection, fu2018learning, qureshi2019adversarial, ghasemipour2019divergence} and} integral probability metrics \cite{kim2018imitation, xiao2019wasserstein}, thereby matching occupancy measures without unnecessary bias.

\begin{figure}[t]
\vspace{-1.5em}
\centering
\makebox[\textwidth][c]{
\subfloat[Classic IRL (Online)]
{\includegraphics[width=0.25625\linewidth]{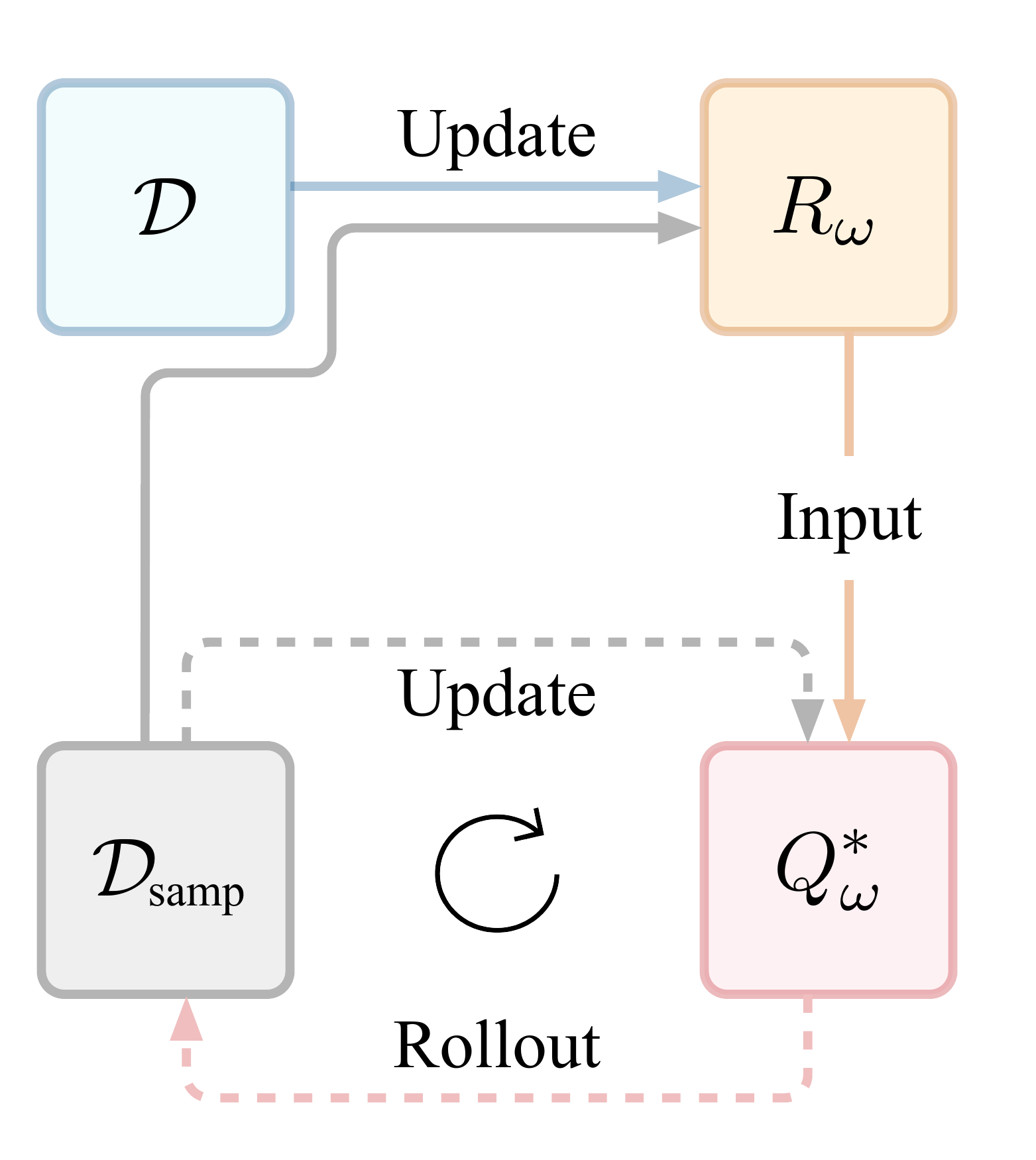}}
\hfill
\subfloat[Adversarial IL (Online)]
{\includegraphics[width=0.25625\linewidth]{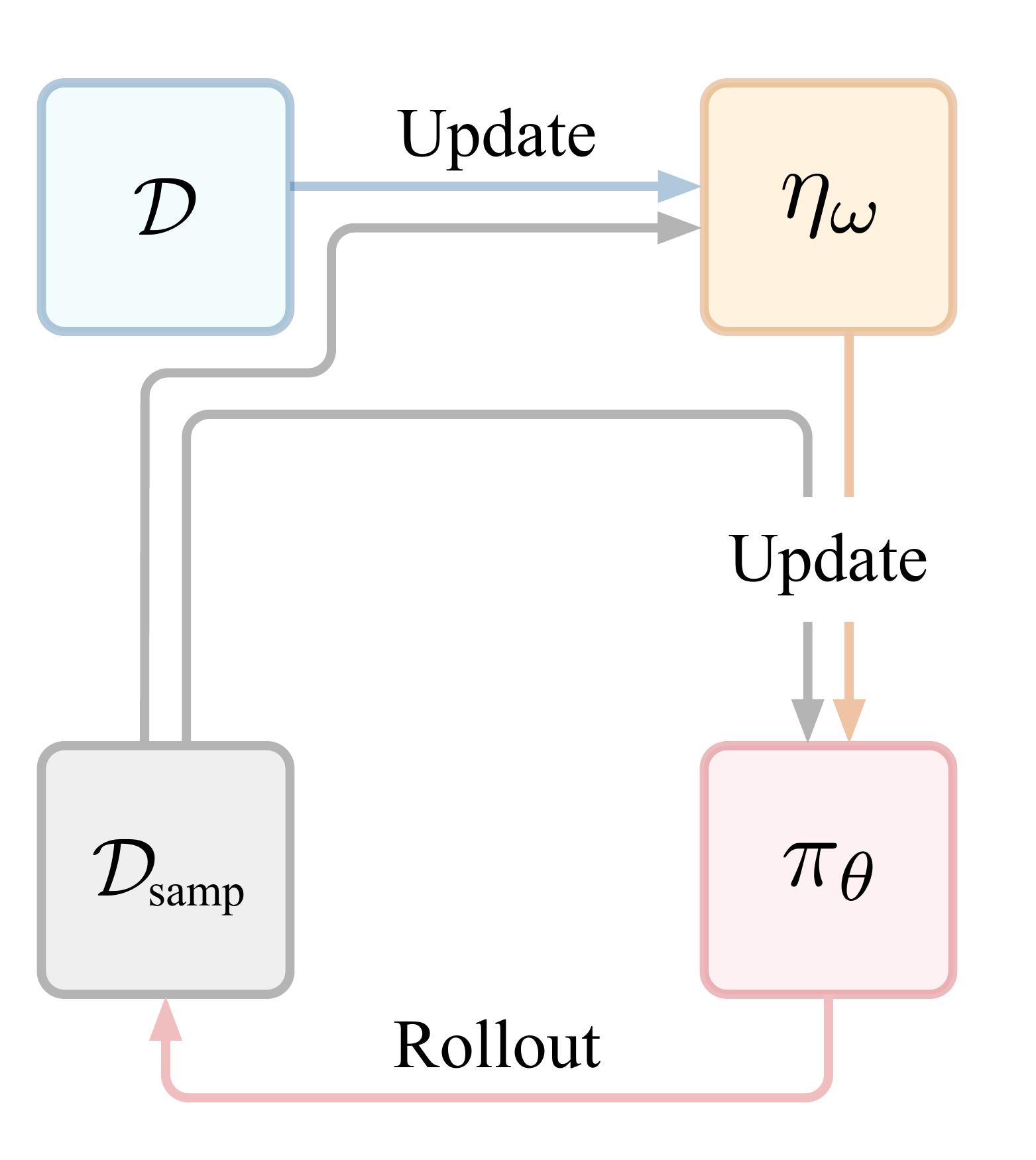}}
\hfill
\subfloat[Off-Policy Adaptations]
{\includegraphics[width=0.25625\linewidth]{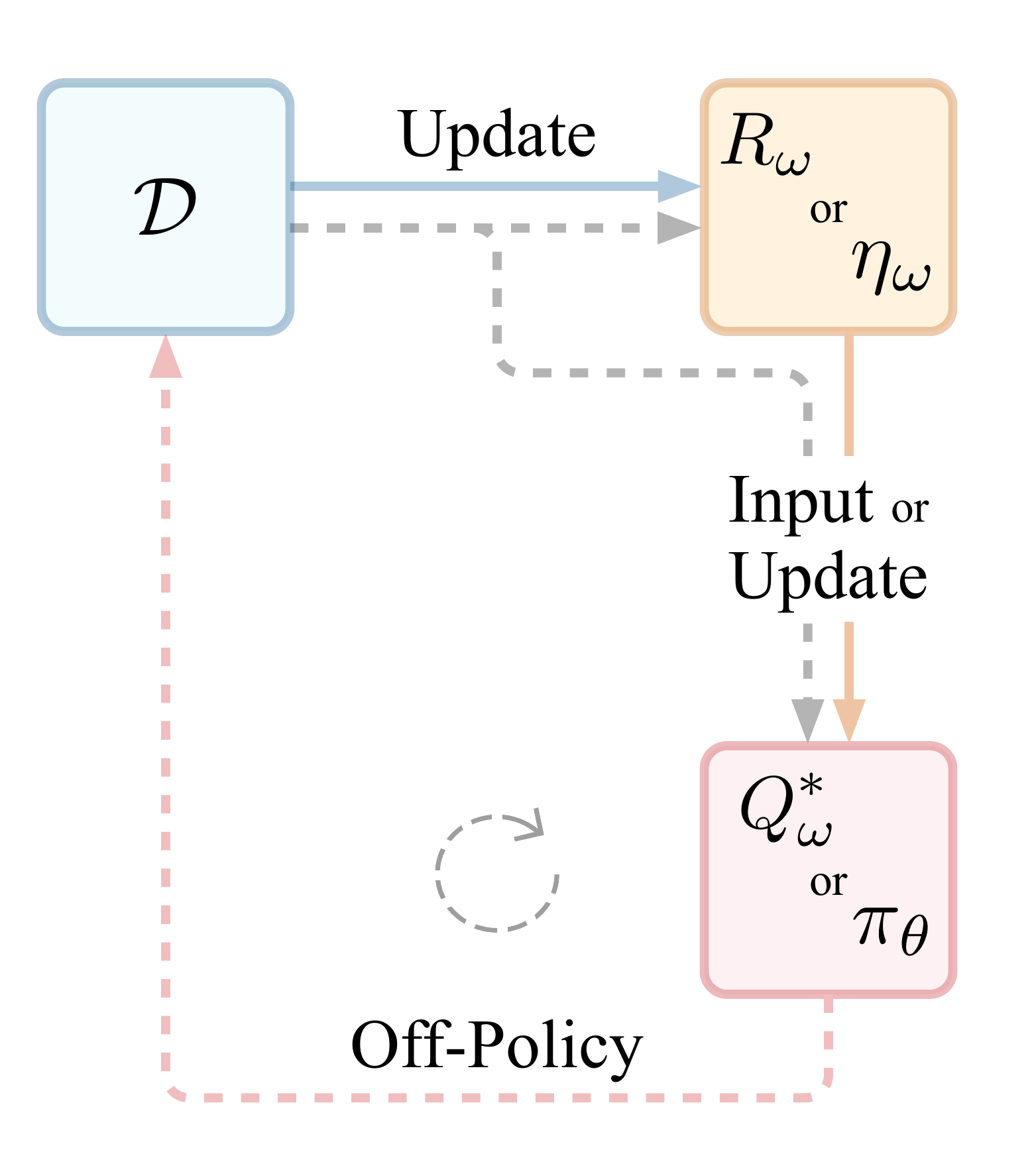}}
\hfill
\subfloat[EDM (Intrinsically Offline)]
{\includegraphics[width=0.25625\linewidth]{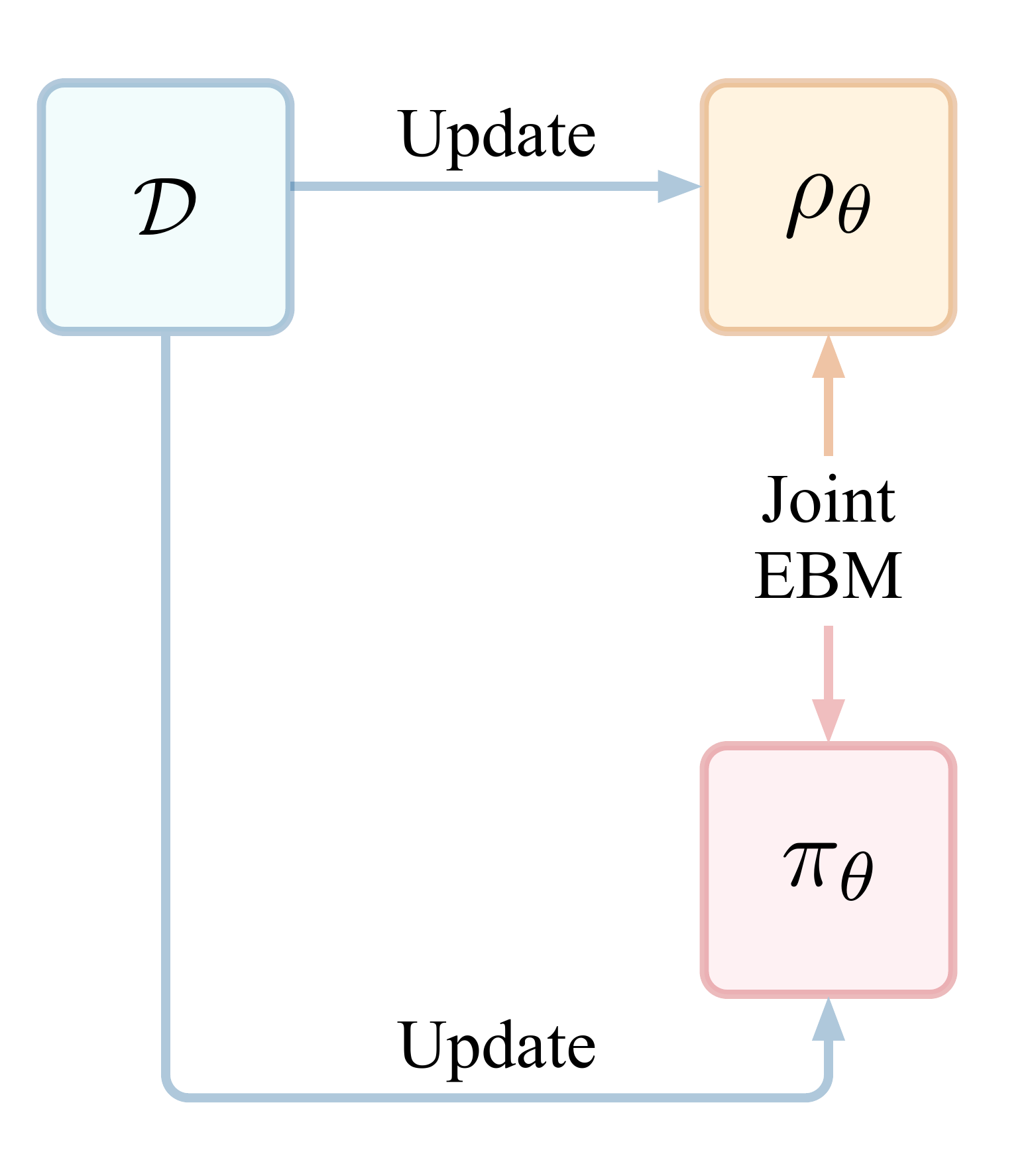}}
}
\vspace{-0.1em}
\caption{\small \dayum{\textit{From Online to Offline Learning}. \textbf{(a)} Classic IRL-based algorithms execute RL repeatedly in an inner loop, learning imitator policies indirectly via parameterizations $\omega$ of a reward function $R_{\omega}$. \textbf{(b)} Adversarial IL methods seek a distribution-matching objective, alternately optimizing a policy $\pi_{\theta}$ parameterized by $\theta$ and a discriminator-like function $\eta_{\omega}$ (which in some cases can be taken as $R$ or a value-function) parameterized by $\omega$. \textbf{(c)} For strictly batch IL, one solution is simply to retrofit existing algorithms from (a) or (b) to work without any sampling actually taking place; this involves using off-policy evaluation as a workaround for these (intrinsically online) apprenticeship methods, which may introduce more variance than desired. \textbf{(d)} We propose a simpler but effective offline method by jointly learning a policy function with an energy-based model of the state distribution.}}
\label{fig:related}
\vspace{-0.9em}
\end{figure}

\textbf{Strictly Batch Imitation Learning}~
Unfortunately, advances in both IRL-based and adversarial IL have a been developed with a very much \textit{online} audience in mind: Precisely, their execution involves repeated on-policy rollouts, which requires access to an environment (for interaction), or at least knowledge of its dynamics (for simulation). Imitation learning in a completely \textit{offline} setting provides neither. On the other hand, while behavioral cloning is ``offline'' to begin with, it is fundamentally limited by disregarding valuable (distributional) information in the demonstration data. Proposed rectifications are infeasible, as they typically require querying the demonstrator, interacting with the environment, or knowledge of model dynamics or sparsity of rewards \cite{syed2007imitation, ross2011reduction, piot2017bridging, attia2018global}. Now of course, an immediate question is whether existing apprenticeship methods can be more-or-less repurposed for batch learning (see Figure \ref{fig:related}). The answer is certainly yes\textemdash but they might not be the most satisfying:

\textit{Adapting Classic IRL}. Briefly, this would inherit the theoretical and computational disadvantages of classic IRL, plus additional difficulties from adapting to batch settings. First, IRL learns imitator policies slowly and indirectly via intermediate parameterizations of $R$, relying on repeated calls to a (possibly imperfect) inner RL procedure. Explicit constraints for tractability also mean that true rewards will likely be imperfectly captured without excessive feature engineering. Most importantly, batch IRL requires \textit{off-policy} evaluation at every step\textemdash which is itself a nontrivial problem with imperfect solutions. For instance, for the max-margin, minimax, and max-likelihood approaches, adaptations for batch imitation \cite{klein2011batch, mori2011model, jain2019model, lee2019truly} rely on least-squares TD and Q-learning, as well as depending on restrictions to linear rewards. Similarly, adaptations of policy-loss and Bayesian IRL in \cite{klein2011batch, tossou2013probabilistic} fall back on linear score-based classification and LSTD. Alternative workarounds involve estimating a model from demonstrations alone \cite{herman2016inverse, tanwani2013inverse}\textemdash feasible only for the smallest or discrete state spaces.

\textit{Adapting Adversarial IL}. Analogously, the difficulty here is that the adversarial formulation requires expectations over trajectories sampled from imitator policy rollouts. Now, there has been recent work focusing on enabling \textit{off-policy} learning through the use of off-policy actor-critic methods \cite{blonde2019sample, kostrikov2019discriminator}. However, this is accomplished by skewing the divergence minimization objective to minimize the distance between the distributions induced by the demonstrator and the replay buffer (instead of the imitator); they must still operate in an online fashion, and are not applicable in a strictly batch setting. More recently, a reformulation in \cite{kostrikov2020imitation} does away with a separate critic by learning the (log density ratio) ``$Q$-function'' via the same objective used for distribution matching. While this theoretically enables fully offline learning, it inherits a similarly complex alternating max-min optimization procedure; moreover, the objective involves the logarithm of an expectation over an exponentiated difference in the Bellman operator\textemdash for which mini-batch approximations of gradients are biased.

\textbf{Three Desiderata}~
At risk of belaboring, the offline setting means that we already have \textit{all} of the information we will ever get, right at the very start. Hanging on to the RL-centric structure of these intrinsically online apprenticeship methods relies entirely on off-policy techniques\textemdash which may introduce more variance than we can afford. In light of the preceding discussion, \mbox{it is clear that a good} solution to the strictly batch imitation learning (SBIL) problem should satisfy the following criteria:
\vspace{-0.5em}
\begin{enumerate}[leftmargin=2em]
\item \dayum{\textbf{Policy}: First, it should directly learn a policy (capturing ``stepwise'' action conditionals) without relying on learning intermediate rewards, and without generic constraints biasing the solution.}
\item \textbf{Occupancy}: But unlike the (purely discriminative) nature of behavioral cloning, it should (generatively) account for information from rollout distributions (capturing ``global'' state marginals).
\item \textbf{Intrinsically Batch}: Finally, it should work offline without known/learned models, and without resorting to off-policy evaluations done within inner loops/max-min optimizations (see Table \ref{tab:related}).
\end{enumerate}

\section{Energy-based Distribution Matching}\label{sec:edm}

We begin by parameterizing with $\theta$ our policy $\pi_{\theta}$, and occupancy measure $\rho_{\theta}$. We are interested in (explicitly) learning a policy while (implicitly) minimizing a divergence between occupancy measures:
\begin{equation}\label{eqn:obj_dm}
\textstyle\argmin_{\theta}D_{\phi}(\rho_{D}\|\rho_{\theta})
\end{equation}
for some choice of generator $\phi$. Note that, unlike in the case of online apprenticeship, our options are significantly constrained by the fact that rollouts of $\pi_{\theta}$ are not actually possible. In the sequel, we shall use $\phi(u)=u\log u$, which gives rise to the (forward) KL, so we write $\argmin_{\theta}D_{\text{KL}}(\rho_{D}\|\rho_{\theta})=\argmin_{\theta}$$-\mathbb{E}_{s,a\sim\rho_{D}}$$\log\rho_{\theta}(s,a)$. Now, consider the general class of stationary policies of the form:
\begin{equation}\label{eqn:policy_class}
\pi_{\theta}(a|s)=\frac{e^{f_{\theta}(s)[a]}}{\sum_{a}e^{f_{\theta}(s)[a]}}
\end{equation}
where $f_{\theta}:\mathcal{S}\rightarrow\mathbb{R}^{\mathcal{A}}$ indicates the logits for action conditionals. An elementary result \cite{feinberg2012handbook, puterman2014markov} shows a bijective mapping between the space of policies and occupancy measures satisfying the Bellman flow constraints,
and $\pi(a|s)$\pix$=$\pix$\rho_{\pi}(s,a)/\rho_{\pi}(s)$; this allows decomposing the log term in the divergence as:
\begin{equation}\label{eqn:policy_identity}
\log\rho_{\theta}(s,a)=\log\rho_{\theta}(s)+\log\pi_{\theta}(a|s)
\end{equation}
\textbf{Objective}~
Ideally, our desired loss is therefore:
\begin{equation}\label{eqn:obj_loss_original}
\mathcal{L}(\theta)=-\mathbb{E}_{s\sim\rho_{D}}\log\rho_{\theta}(s)-\mathbb{E}_{s,a\sim\rho_{D}}\log\pi_{\theta}(a|s)
\end{equation}
with the corresponding gradient given by:
\begin{equation}\label{eqn:obj_loss_gradient}
\nabla_{\theta}\mathcal{L}(\theta)=-\mathbb{E}_{s\sim\rho_{D}}\nabla_{\theta}\log\rho_{\theta}(s)-\mathbb{E}_{s,a\sim\rho_{D}}\nabla_{\theta}\log\pi_{\theta}(a|s)
\end{equation}
Now, there is an obvious problem. Backpropagating through the first term is impossible as we cannot compute $\rho_{\theta}(s)$\textemdash nor do we have access to online rollouts of $\pi_{\theta}$ to explicitly estimate it. In this offline imitation setting, our goal is to answer the question: Is there any benefit in \textit{learning} an approximate model in its place instead? Here we consider energy-based modeling \cite{lecun2006tutorial}, which associates scalar measures of compatibility (i.e. energies) with configurations of variables (i.e. states). Specifically, we take advantage of the \textit{joint energy-based modeling} approach \cite{grathwohl2020your, du2019implicit, xie2016theory}\textemdash in particular the proposal for a classifier to be simultaneously learned with a density model defined implicitly by the logits of the classifier (which\textemdash as they observe\textemdash yields improvements such as in calibration and robustness):

\textbf{Joint Energy-based Modeling}~
Consider first the general class of energy-based models (EBMs) for state occupancy measures \smash{$\rho_{\theta}(s)\propto$}\pix\raisebox{-1pt}{\smash{$e^{-E(s)}$}}. Now, mirroring the exposition in \cite{grathwohl2020your}, note that a model of the state-action occupancy measure \smash{$\rho_{\theta}(s,a)=e^{f_{\theta}(s)[a]}/Z_{\theta}$} can be defined via the parameterization for $\pi_{\theta}$, where \smash{$Z_{\theta}$} is the partition function. The state-only model for \smash{$\rho_{\theta}(s)=\sum_{a}e^{f_{\theta}(s)[a]}/Z_{\theta}$} is then obtained by marginalizing out $a$. In other words, the parameterization of $\pi_{\theta}$ already implicitly defines an EBM of state visitation distributions with the energy function \smash{$E_{\theta}:$}\pix\raisebox{-1pt}{\smash{$\mathbb{R}^{|\mathcal{S}|}\rightarrow\mathbb{R}^{|\mathcal{A}|}$}} given as follows:
\vspace{-0.5em}
\begin{equation}\label{eqn:energy}
E_{\theta}(s)\doteq-\log\textstyle\sum_{a}e^{f_{\theta}(s)[a]}
\end{equation}
\dayum{The chief difference from \cite{grathwohl2020your}, of course, is that here the true probabilities in question are not static class conditionals/marginals: The \textit{actual} occupancy measure corresponds to rolling out $\pi_{\theta}$, and if we could do that, we would naturally recover an approach not unlike the variety of distribution-aware algorithms in the literature; see e.g. \cite{schroecker2017state}. In the strictly batch setting, we clearly cannot sample directly from this (online) distribution. However, as a matter of multitask learning, we still hope to gain from jointly learning an (offline) \textit{model} of the state distribution---which we can then freely sample from:}

\begin{reproposition}[restate=surrogate,name=Surrogate Objective]\upshape\label{thm:surrogate}
Define the ``occupancy'' loss $\mathcal{L}_{\rho}$ as the difference in energy:
\begin{equation}
\mathcal{L}_{\rho}(\theta)\doteq\mathbb{E}_{s\sim\rho_{D}}E_{\theta}(s)-\mathbb{E}_{s\sim\rho_{\theta}}E_{\theta}(s) \label{eqn:obj_loss_rho}
\end{equation}
Then \smash{$\nabla_{\theta}\mathcal{L}_{\rho}(\theta)=-\mathbb{E}_{s\sim\rho_{D}}\nabla_{\theta}\log\rho_{\theta}(s)$}. In other words, differentiating this recovers the first term in Equation \ref{eqn:obj_loss_gradient}. Therefore if we define a standard ``policy'' loss \smash{$\mathcal{L}_{\pi}(\theta)\doteq-\mathbb{E}_{s,a\sim\rho_{D}}\log\pi_{\theta}(a|s)$}, then:
\begin{equation}\label{eqn:obj_loss_surrogate}
\mathcal{L}_{\text{surr}}(\theta)\doteq\mathcal{L}_{\rho}(\theta)+\mathcal{L}_{\pi}(\theta)
\end{equation}
yields a surrogate objective that can be optimized, instead of the original $\mathcal{L}$. Note that by relying on the offline energy-based model, we now have access to the gradients of the terms in the expectations.
\end{reproposition}

\begin{algorithm}[t]
\begin{spacing}{1.0}
\begin{algorithmic}[1]
\STATE\textbf{Input}: SGLD hyperparameters $\alpha,\sigma$, PCD hyperparameters $\kappa,\iota,\delta$, and mini-batch size $N$
\STATE\textbf{Initialize}: Policy network parameters $\theta$, and PCD buffer $B_{\kappa}$
\WHILE{not converged}
\STATE Sample \smash{$(s_{1},a_{1}),...,(s_{N},a_{N})\sim\mathcal{D}$} from demonstrations dataset
\STATE Sample \smash{$(\tilde{s}_{1,0},...,\tilde{s}_{N,0})$ as $\tilde{s}_{n,0}\sim B_{\kappa}$} \textbf{w.p.} $1-\delta$ \textbf{o.w.} \smash{$\tilde{s}_{n,0}\sim\mathcal{U}(\mathcal{S})$}
\FOR{$i=1,...,\iota$}
\STATE \raisebox{2pt}{\smash{$\tilde{s}_{n,i}=\tilde{s}_{n,i-1}-\alpha\cdot\partial E_{\theta}(\tilde{s}_{n,i-1})/\partial\tilde{s}_{n,i-1}+\sigma\cdot\mathcal{N}(0,I)$, $\forall n\in\{1,...,N\}$}}
\ENDFOR
\STATE \raisebox{0pt}{\smash{$\hphantom{oo\pix\pix\pix}\mathllap{\hat{\mathcal{L}}_{\pi}}\gets\frac{1}{N}\sum_{n=1}^{N}\text{CrossEntropy}(\pi_{\theta}(\cdot|s_n),a_n)$}}
\COM{$\mathcal{L}_{\pi}=-\mathbb{E}_{s,a\sim\rho_{D}}\log\pi_{\theta}(a|s)$}
\STATE \raisebox{-1.5pt}{\smash{$\hphantom{oo\pix\pix\pix}\mathllap{\hat{\mathcal{L}}_{\rho}}\gets\frac{1}{N}\sum_{n=1}^{N}E_{\theta}(s_{n})-\frac{1}{N}\sum_{n=1}^{N}E_{\theta}(\tilde{s}_{n,\iota})$}}
\COM{$\mathcal{L}_{\rho}=\mathbb{E}_{s\sim\rho_{D}}E_{\theta}(s)-\mathbb{E}_{s\sim\rho_{\theta}}E_{\theta}(s)$}
\STATE \raisebox{-2pt}{Add \smash{$\tilde{s}_{n,\iota}$ to $B_{\kappa}$, $\forall n\in\{1,...,N\}$}}
\STATE \raisebox{-2pt}{Backpropagate \smash{$\nabla_{\theta}\hat{\mathcal{L}}_{\rho}+\nabla_{\theta}\hat{\mathcal{L}}_{\pi}$}}
\ENDWHILE
\STATE\textbf{Output}: Learned policy parameters $\theta$
\end{algorithmic}
\end{spacing}
\caption{Energy-based Distribution Matching \COM{for Strictly Batch Imitation Learning}}\label{alg:edm}
\end{algorithm}

\vspace*{-20pt}

\textit{Proof}. Appendix \ref{app:proofs}. Sketch: For any $s$, write \smash{$\rho_{\theta}(s)=e^{-E_{\theta}(s)}/\int_{\mathcal{S}}e^{-E_{\theta}(s)}ds$}, for which the gradient of the logarithm is given by \smash{$-\nabla_{\theta}\log\rho_{\theta}(s)=\nabla_{\theta}E_{\theta}(s)-\mathbb{E}_{s\sim\rho_{\theta}}\nabla_{\theta} E_{\theta}(s)$}. Then, taking expectations over $\rho_{D}$ and substituting in the energy term as given by Equation \ref{eqn:energy}, straightforward manipulation shows $-\nabla_{\theta}\mathbb{E}_{s\sim\rho_{D}}\log\rho_{\theta}(s)=\nabla_{\theta}\mathcal{L}_{\rho}(\theta)$. The second part follows immediately from Equation \ref{eqn:obj_loss_original}.
\QED

\dayum{Why is this better than before? Because using the original objective $\mathcal{L}$ required us to know $\rho_{\theta}(s)$, which---even modeled separately as an EBM---we do not (since we cannot compute the normalizing constant). On the other hand, using the surrogate objective $\mathcal{L}_{\text{surr}}$ only requires being able to sample from the EBM, which is easier. Note that jointly learning the EBM does not constrain/bias the policy, as this simply reuses the policy parameters along with the extra degree of freedom in the logits $f_{\theta}(s)[\cdot]$.}

\textbf{Optimization}~
The EDM surrogate objective entails minimal addition to the standard behavioral cloning loss. Accordingly, it is perfectly amenable to mini-batch gradient approximations\textemdash unlike for instance \cite{kostrikov2020imitation}, for which mini-batch gradients are biased in general. We approximate the expectation over $\rho_{\theta}$ in Equation \ref{eqn:obj_loss_rho} via stochastic gradient Langevin dynamics (SGLD) \cite{welling2011bayesian}, which follows recent successes in training EBMs parameterized by deep networks \cite{du2019implicit, grathwohl2020your, nijkamp2020on}, and use persistent contrastive divergence (PCD) \cite{tieleman2008training} for computational savings. Specifically, each sample is drawn as:
\begin{equation}
\tilde{s}_{i}=\tilde{s}_{i-1}-\alpha\cdot\frac{\partial E_{\theta}(\tilde{s}_{i-1})}{\partial\tilde{s}_{i-1}}+\sigma\cdot\mathcal{N}(0,I)
\end{equation}
\dayum{where $\alpha$ denotes the SGLD learning rate, and $\sigma$ the noise coefficient. Algorithm \ref{alg:edm} details the EDM optimization procedure, with a buffer $B_{\kappa}$ of size $\kappa$, reinitialization frequency $\delta$, and number of iterations $\iota$, where $\tilde{s}_{0}\sim\rho_{0}(s)$ is sampled uniformly. Note that the buffer here should not be confused with the ``replay buffer'' within (online) imitation learning algorithms, to which it bears no relationship whatsoever. In practice, we find that the configuration given in \cite{grathwohl2020your} works effectively with only small modifications. We refer to \cite{grathwohl2020your, du2019implicit, xie2016theory, lecun2006tutorial, welling2011bayesian, tieleman2008training} for discussion of general considerations for EBM optimization.}

\begin{table}[t]\small
\newcolumntype{O}{>{          \arraybackslash}m{0.4 cm}}
\newcolumntype{A}{>{          \arraybackslash}m{2.8 cm}}
\newcolumntype{B}{>{\centering\arraybackslash}m{1.85cm}}
\newcolumntype{C}{>{\centering\arraybackslash}m{1.9 cm}}
\newcolumntype{D}{>{\centering\arraybackslash}m{2.1 cm}}
\newcolumntype{E}{>{\centering\arraybackslash}m{1.8 cm}}
\newcolumntype{F}{>{\centering\arraybackslash}m{2.1 cm}}
\newcolumntype{G}{>{\centering\arraybackslash}m{1.7 cm}}
\setlength\tabcolsep{0pt}
\renewcommand{\arraystretch}{0.93}
\vspace{-1.5em}
\caption{\dayum{\textit{From Online to Offline Imitation}. Recall the three desiderata from Section \ref{sec:sbil}, where we seek an SBIL solution that: \textbf{(1)} learns a \textit{directly parameterized} policy, without restrictive constraints biasing the solution\textemdash e.g. restrictions to linear/convex function classes for intermediate rewards, or generic norm-based penalties on reward sparsity; \textbf{(2)} is \textit{dynamics-aware} by accounting for distributional information\textemdash either through temporal or parameter consistency; and \textbf{(3)} is \textit{intrinsically batch}, in the sense of being operable strictly offline, and directly optimizable\textemdash i.e. without recourse to off-policy evaluations in costly inner loops or alternating max-min optimizations.
}}
\label{tab:related}
\begin{center}
\begin{adjustbox}{max width=\textwidth}
\input{table/related_work}
\end{adjustbox}
\end{center}
\vspace{-1.75em}
\end{table}

\vspace{-0.5em}
\section{Analysis and Interpretation}\label{sec:discussion}
\vspace{-0.3em}

Our development in Section \ref{sec:edm} proceeded in three steps. First, we set out with a divergence minimization objective in mind (Equation \ref{eqn:obj_dm}). With the aid of the decomposition in Equation \ref{eqn:policy_identity}, we obtained the original (online) maximum-likelihood objective function (Equation \ref{eqn:obj_loss_original}). Finally, using Proposition \ref{thm:surrogate}, we instead optimize an (offline) joint energy-based model by scaling the gradient of a surrogate objective (Equation \ref{eqn:obj_loss_surrogate}). Now, the mechanics of the optimization are straightforward, but what is the underlying motivation for doing so? In particular, how does the EDM objective relate to existing notions of (1) divergence minimization, (2) joint learning, as well as (3) imitation learning in general?

\dayum{
\textbf{Divergence Minimization}~
With the seminal observation by \cite{ho2016generative} of the equivalence in Equation \ref{eqn:obj_rl_irl}, the IL arena was quickly populated with a lineup of adversarial algorithms minimizing a variety of distances \cite{fu2018learning, qureshi2019adversarial, ghasemipour2019divergence, kim2018imitation, xiao2019wasserstein}, and the forward KL in this framework was first investigated in \cite{ghasemipour2019divergence}. However in the strictly batch setting, we have no ability to compute (or even sample from) the actual rollout distribution for $\pi_{\theta}$, so we instead choose to learn an EBM in its place. To be clear, we are now doing something quite different than \cite{fu2018learning, qureshi2019adversarial, ghasemipour2019divergence, kim2018imitation, xiao2019wasserstein}: In minimizing the divergence (Equation \ref{eqn:obj_dm}) by simultaneously learning an (offline) model instead of sampling from (online) rollouts, $\pi_{\theta}$ and $\rho_{\theta}$ are no longer coupled in terms of rollout \textit{dynamics}, and the coupling that remains is in terms of the underlying \textit{parameterization} $\theta$. That is the price we pay. At the same time, hanging on to the adversarial setup in the batch setting requires estimating intrinsically on-policy terms via off-policy methods, which are prone to suffer from high variance. Moreover, the divergence minimization interpretations of adversarial IL hinge crucially on the assumption that the discriminator-like function is perfectly optimized \cite{ho2016generative, fu2018learning, ghasemipour2019divergence, kostrikov2020imitation}\textemdash which may not be realized in practice offline. The EDM objective aims to sidestep both of these difficulties.}

\textbf{Joint Learning}~
In the online setting, minimizing Equation \ref{eqn:obj_loss_original} is equivalent to injecting temporal consistency into behavioral cloning: While the \smash{$\mathbb{E}_{s,a\sim\rho_{D}}\log\pi_{\theta}(a|s)$} term is purely a discriminative objective, the \smash{$\mathbb{E}_{s\sim\rho_{D}}\log\rho_{\theta}(s)$} term additionally constrains \smash{$\pi_{\theta}(\cdot|s)$} to the space of policies for which the induced state distribution matches the data. In the offline setting, instead of this \textit{temporal} relationship we are now leveraging the \textit{parameter} relationship between $\pi_{\theta}$ and $\rho_{\theta}$\textemdash that is, from the joint EBM. In effect, this accomplishes an objective similar to multitask learning, where representations of both discriminative (policy) and generative (visitation) distributions are learned by sharing the same underlying function approximator. As such, (details of sampling techniques aside) this additional mandate does \textit{not} add any bias. This is in contrast to generic approaches to regularization in IL, such as the norm-based penalties on the sparsity of implied rewards \cite{piot2014boosted, piot2017bridging, reddy2020sqil}\textemdash which adds bias. The state-occupancy constraint in EDM simply harnesses the extra degree of freedom hidden in the logits \smash{$f_{\theta}(s)$}\textemdash which are normally allowed to shift by an arbitrary scalar\textemdash to define the density over states.

\textbf{Imitation Learning}~
Finally, recall the classical notion of \textit{imitation learning} that we started with (Equation \ref{eqn:obj_il}). As noted earlier, naive application by behavioral cloning simply ignores the endogeneity of the rollout distribution. How does our final surrogate objective (Equation \ref{eqn:obj_loss_surrogate}) relate to this? First, we place Equation \ref{eqn:obj_il} in the maximum entropy RL framework in order to speak in a unified language:

\begin{reproposition}[restate=classical,name=Classical Objective]\upshape\label{thm:classical}
Consider the classical IL objective in Equation \ref{eqn:obj_il}, with policies parameterized as Equation \ref{eqn:policy_class}. Choosing $\mathcal{L}$ to be the (forward) KL divergence yields the following:
\begin{equation}
\textstyle\argmax_{R}\big(\mathbb{E}_{s\sim\rho^{*}_{R}}\mathbb{E}_{a\sim\pi_{D}(\cdot|s)}Q^{*}_{R}(s,a)-\mathbb{E}_{s\sim\rho^{*}_{R}}V^{*}_{R}(s)\big) \label{eqn:obj_il_classic}
\end{equation}
where $Q^{*}_{R}:\mathcal{S}\times\mathcal{A}\rightarrow\mathbb{R}$ is the (soft) $Q$-function given by $Q^{*}_{R}(s,a)=R(s,a)+\gamma\mathbb{E}_{T}[V^{*}_{R}(s^{\prime})|s,a]$, $V^{*}(s)\in\mathbb{R}^{\mathcal{S}}$ is the (soft) value function $V^{*}_{R}(s)=\log\sum_{a}e^{Q^{*}_{R}(s,a)}$, and $\rho^{*}_{R}$ is the occupancy for $\pi^{*}_{R}$.
\end{reproposition}

\textit{Proof}. Appendix \ref{app:proofs}. This relies on the fact that we are free to identify the logits $f_{\theta}$ of our policy with a (soft) $Q$-function. Specifically, this requires the additional fact that the mapping between $Q$-functions and reward functions is bijective, which we also state (and prove) as Lemma \ref{thm:biject} in Appendix \ref{app:proofs}.
\QED

\dayum{
This is intuitive: It states that classical imitation learning with $\mathcal{L}=D_{\text{KL}}$ is equivalent to searching for a reward function $R$. In particular, we are looking for an $R$ for which\textemdash in expectation over \textit{rollouts} of policy $\pi^{*}_{R}$\textemdash the advantage function $Q^{*}_{R}(s,a)-V^{*}_{R}(s)$ for taking actions $a\sim\pi_{D}(\cdot|s)$ is maximal. Now, the following distinction is key: While Equation \ref{eqn:obj_il_classic} is perfectly valid as a choice of objective, it is a certain (naive) substitution in the offline setting that is undesirable. Specifically, Equation \ref{eqn:obj_il_classic} is precisely what behavioral cloning attempts to do, but\textemdash without the ability to perform $\pi^{*}_{R}$ rollouts\textemdash it simply replaces \smash{$\rho^{*}_{R}$} with \smash{$\rho_{D}$}. This is \textit{not} an (unbiased) ``approximation'' in, say, the sense that \smash{$\hat{\mathcal{L}}_{\rho}$} empirically approximates $\mathcal{L}_{\rho}$, and is especially inappropriate when $\rho_{D}$ contains very few demonstrations to begin with. While EDM cannot fully ``undo'' the damage (nothing can do that in the strictly batch setting), it uses a ``smoothed'' EBM in place of $\rho_{D}$, which---as we shall see empirically---leads to largest improvements precisely when the number of demonstrations are few.}


\begin{reproposition}[restate=bcedm,name=From BC to EDM]\upshape\label{thm:bcedm}
The behavioral cloning objective is equivalently the following, where\textemdash compared to Equation \ref{eqn:obj_il_classic}\textemdash expectations over states are now taken w.r.t. $\rho_{D}$ instead of $\rho^{*}_{R}$:
\begin{equation}
\setlength{\abovedisplayskip}{4pt}
\setlength{\belowdisplayskip}{4pt}
\textstyle\argmax_{R}\big(\mathbb{E}_{s\sim\rho_{D}}\mathbb{E}_{a\sim\pi_{D}(\cdot|s)}Q^{*}_{R}(s,a)-\mathbb{E}_{s\sim\rho_{D}}V^{*}_{R}(s)\big) \label{eqn:obj_il_bc}
\end{equation}
In contrast, by augmenting the (behavioral cloning) ``policy'' loss $\mathcal{L}_{\pi}$ with the ``occupancy'' loss $\mathcal{L}_{\rho}$, what the EDM surrogate objective achieves is to replace one of the expectations with the learned $\rho_{\theta}$:
\begin{equation} \label{eqn:obj_il_edm}
\setlength{\abovedisplayskip}{4pt}
\setlength{\belowdisplayskip}{4pt}
\textstyle\argmax_{R}\big(\mathbb{E}_{s\sim\rho_{D}}\mathbb{E}_{a\sim\pi_{D}(\cdot|s)}Q^{*}_{R}(s,a)-\mathbb{E}_{s\sim\rho_{\theta}}V^{*}_{R}(s)\big)
\end{equation}
\end{reproposition}
\vspace{-0.5em}
\textit{Proof}. Appendix \ref{app:proofs}. The reasoning for both statements follows a similar form as for Proposition \ref{thm:classical}.
\QED

Note that by swapping out $\rho^{*}_{R}$ for $\rho_{D}$ in behavioral cloning, the (dynamics) relationship between $\pi^{*}_{R}$ and its induced occupancy measure is (completely) broken, and the optimization in Equation \ref{eqn:obj_il_bc} is equivalent to performing a sort of inverse reinforcement learning with no constraints whatsoever on $R$. What the EDM surrogate objective does is to ``repair'' one of the expectations to allow sampling from a smoother model distribution $\rho_{\theta}$ than the (possibly very sparse) data distribution $\rho_{D}$. (Can we also ``repair'' the other term? But this is now asking to somehow warp $\mathbb{E}_{s\sim\rho_{D}}\mathbb{E}_{a\sim\pi_{D}(\cdot|s)}Q^{*}_{R}(s,a)$ into \smash{$\mathbb{E}_{s\sim\rho_{\theta}}\mathbb{E}_{a\sim\pi_{D}(\cdot|s)}Q^{*}_{R}(s,a)$}. All else equal, this is certainly impossible without querying the expert.)

\vspace{-0.5em}
\section{Experiments}\label{sec:experiments}
\vspace{-0.3em}

\textbf{Benchmarks}~
We test Algorithm \ref{alg:edm} (\textbf{EDM}) against the following benchmarks, varying the amount of demonstration data $\mathcal{D}$ (from a single trajectory to 15) to illustrate sample complexity: The intrinsically offline behavioral cloning (\textbf{BC}), and reward-regularized classification (\textbf{RCAL}) \cite{piot2017bridging}\textemdash which proposes to leverage dynamics information through a sparsity-based penalty on the implied rewards; the deep successor feature network (\textbf{DFSN}) algorithm of \cite{lee2019truly}\textemdash which is an off-policy adaptation of the max-margin IRL algorithm and a (deep) generalization of earlier (linear) approaches by LSTD \cite{klein2011batch, tossou2013probabilistic}; and the state-of-the-art in sample-efficient adversarial imitation learning (\textbf{VDICE}) in \cite{kostrikov2020imitation}, which\textemdash while designed with an online audience in mind\textemdash can theoretically operate in a completely offline manner. (Remaining candidates in Table \ref{tab:related} are inapplicable, since they either only operate in discrete states \cite{herman2016inverse, jain2019model}, or only output a reward \cite{klein2012inverse}, which\textemdash in the strictly batch setting\textemdash does not yield a policy.

\textbf{Demonstrations}~
We conduct experiments on control tasks and a real-world healthcare dataset. For the former, we use OpenAI gym environments \cite{brockman2016openai} of varying complexity from standard RL literature: \texttt{CartPole}, which balances a pendulum on a frictionless track \cite{barto1983neuronlike}, \texttt{Acrobot}, which swings a system of joints up to a given height \cite{geramifard2015rlpy}, \texttt{BeamRider}, which controls an Atari 2600 arcade space shooter \cite{bellemare13arcade}, as well as \texttt{LunarLander}, which optimizes a rocket trajectory for successful landing \cite{klimov2019lunar}. Demonstration datasets $\mathcal{D}$ are generated using pre-trained and hyperparameter-optimized agents from the RL Baselines Zoo \cite{raffin2018rl} in Stable OpenAI Baselines \cite{hill2018stable}. For the healthcare application, we use \texttt{MIMIC-III}, a real-world medical dataset consisting of patients treated in intensive care units from the \href{https://mimic.physionet.org}{Medical Information Mart for Intensive Care} \cite{johnson2016mimic}, which records trajectories of physiological states and treatment actions (e.g. antibiotics and ventilator support) for patients at one-day intervals.

\textbf{Implementation}~
The experiment is arranged as follows: Demonstrations $\mathcal{D}$ are sampled for use as input to train all algorithms, which are then evaluated using 300 live episodes (for OpenAI gym environments) or using a held-out test set (for \texttt{MIMIC-III}). This process is then repeated for a total 50 times (using different $\mathcal{D}$ and randomly initialized networks), from which we compile the means of the performances (and their standard errors) for each algorithm. Policies trained by all algorithms share the same network architecture: two hidden layers of 64 units each with ELU activation (or\textemdash for Atari\textemdash three convolutional layers with ReLU activation). For DSFN, we use the publicly available source code at \cite{lee2019github}, and likewise for VDICE, which is available at \cite{kostrikov2020github}. Note that VDICE is originally designed for Gaussian actions, so we replace the output layer of the actor with a Gumbel-softmax parameterization; offline learning is enabled by setting the ``replay regularization'' coefficient to zero. Algorithm \ref{alg:edm} is implemented using the source code for joint EBMs \cite{grathwohl2020your} publicly available at \cite{grathwohl2020github}, which already contains an implementation of SGLD. Note that the only difference between BC and EDM is the addition of $\mathcal{L}_{\rho}$, and the RCAL loss is straightforwardly obtained by inverting the Bellman equation. See Appendix \ref{app:more_experiments} for additional detail on experiment setup, benchmarks, and environments.

\begin{figure}[t]
\vspace{-1.5em}
\centering
\makebox[\textwidth][c]{
\subfloat[\texttt{Acrobot}]{
\adjustbox{trim={0.02\width} {0.05\height} {0.05\width} {0.0\height},clip}{
\includegraphics[width=0.25625\linewidth, trim=8em 0em 7em 15em]{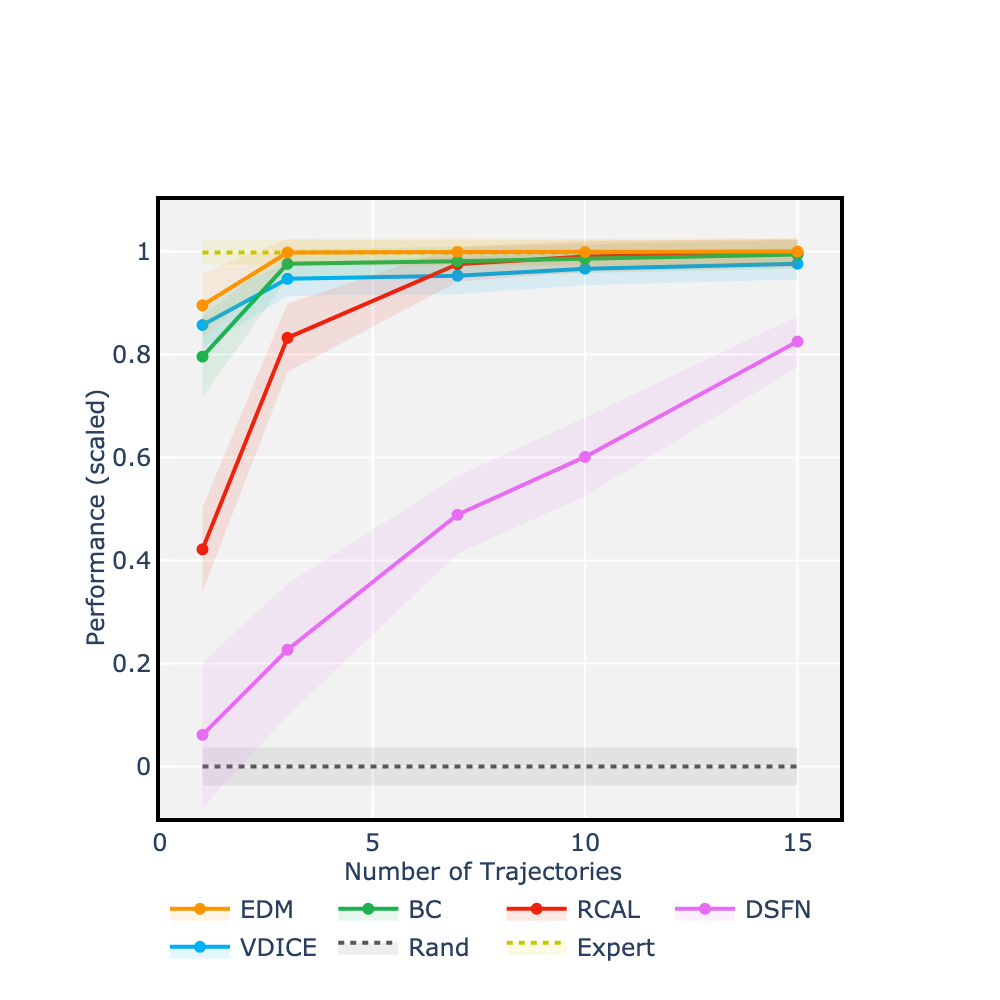}}
}
\hfill
\subfloat[\texttt{CartPole}]{
\adjustbox{trim={0.02\width} {0.05\height} {0.05\width} {0.0\height},clip}{
\includegraphics[width=0.25625\linewidth, trim=8em 0em 7em 15em]{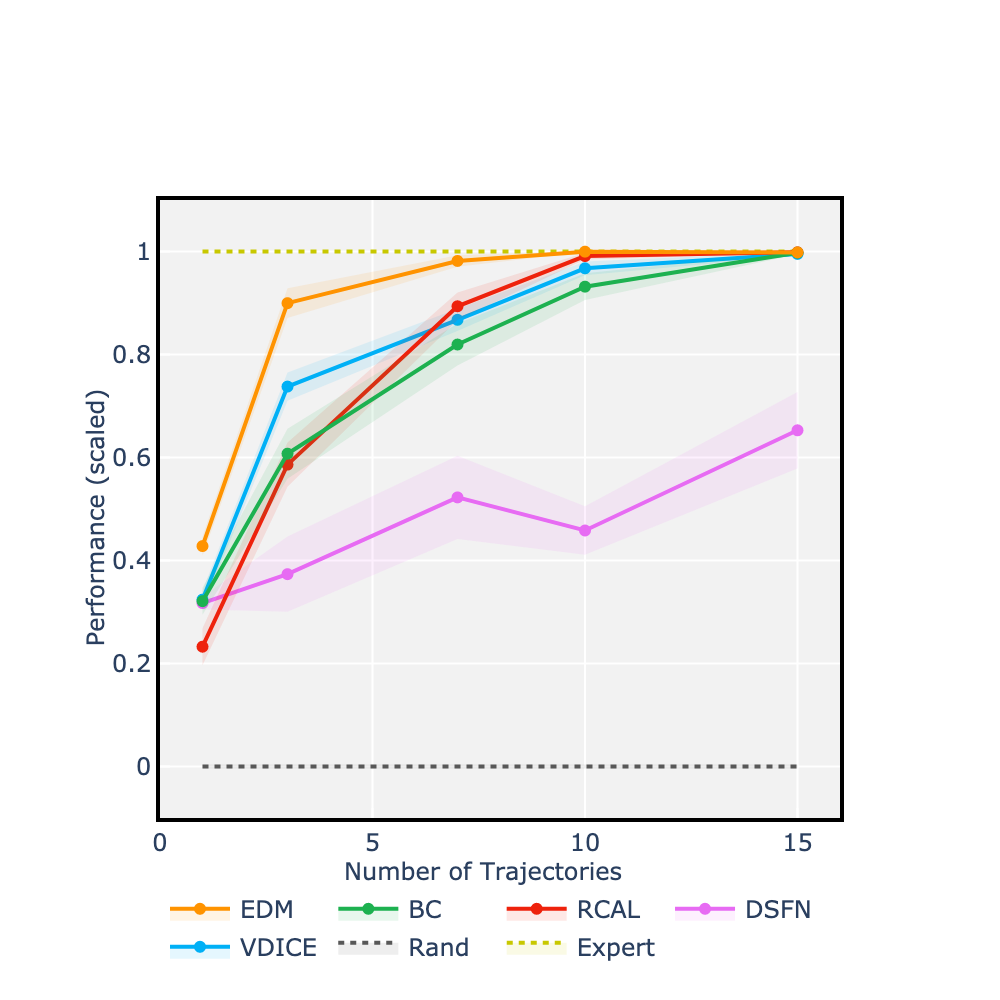}}
}
\hfill
\subfloat[\texttt{LunarLander}]{
\adjustbox{trim={0.02\width} {0.05\height} {0.05\width} {0.0\height},clip}{
\includegraphics[width=0.25625\linewidth, trim=8em 0em 7em 15em]{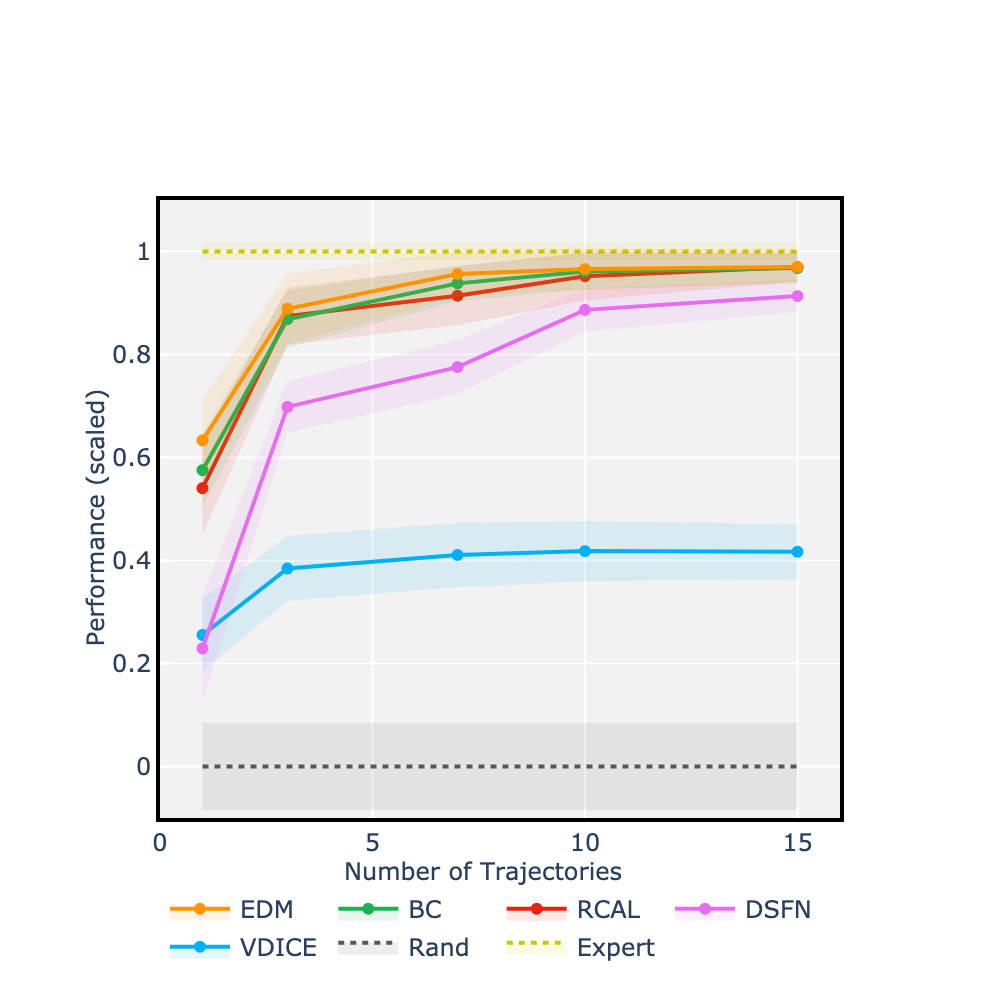}}
}
\hfill
\subfloat[\texttt{BeamRider}]{
\adjustbox{trim={0.02\width} {0.05\height} {0.05\width} {0.0\height},clip}{
\includegraphics[width=0.25625\linewidth, trim=8em 0em 7em 15em]{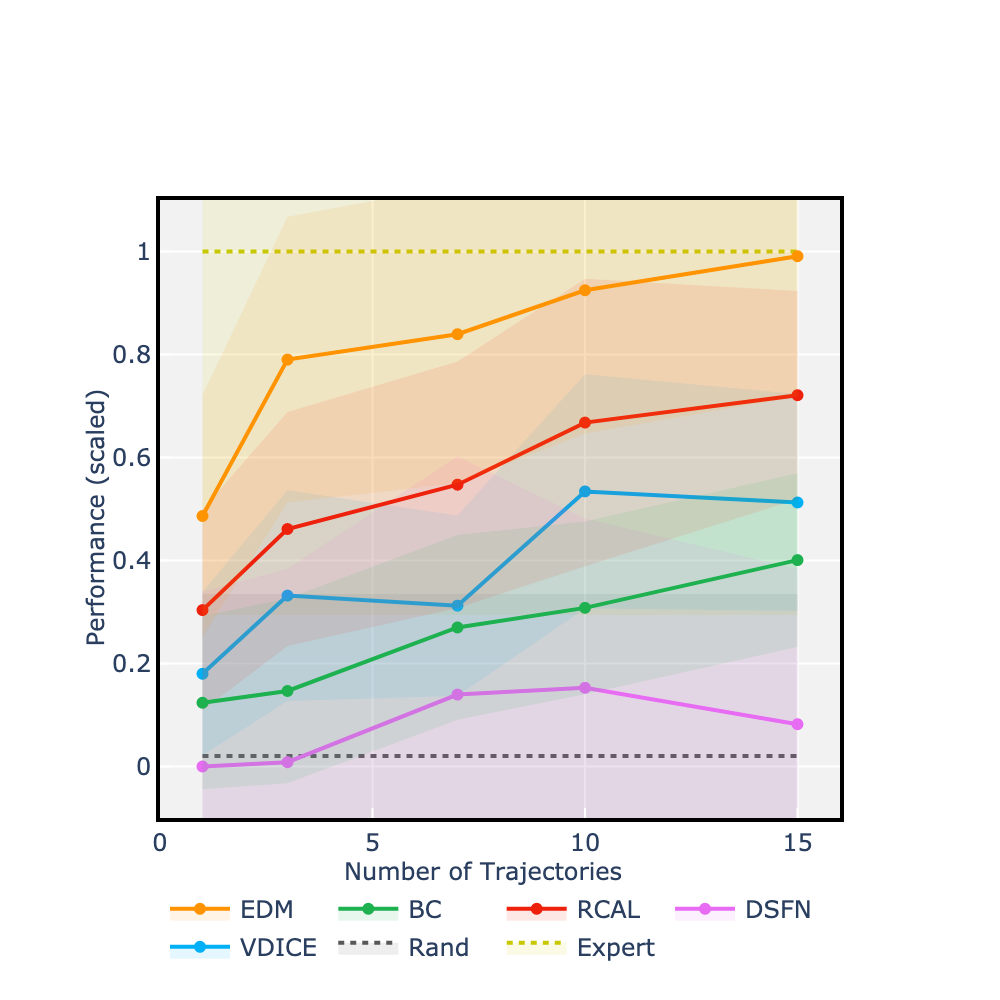}}
}~
}
\vspace{-0.5em}
\caption{\small \textit{Performance Comparison for Gym Environments}. The $x$-axis indicates the amount of demonstration data provided (i.e. number of trajectories, in $\{$1,\pix3,\pix7,\pix10,\pix15$\}$), and the $y$-axis shows the average returns of each imitation algorithm (scaled so that the demonstrator attains a return of 1 and a random policy network attains 0).}
\label{fig:results}
\vspace{-0.75em}
\end{figure}

\textbf{Evaluation and Results}~
For gym environments, the performance of trained imitator policies (learned offline) is evaluated with respect to (true) \textit{average returns} generated by deploying them live. Figure \ref{fig:results} shows the results for policies given different numbers of trajectories as input to training, and Appendix \ref{app:more_experiments} provides exact numbers. For the \texttt{MIMIC-III} dataset, policies are trained and tested on demonstrations by way of cross-validation; since we have no access to ground-truth rewards, we assess performance according to \textit{action-matching} on held-out test trajectories, per standard \cite{lee2019github}; Table \ref{tab:mimic} shows the results. With respect to either metric, we find that EDM consistently produces policies that perform similarly or better than benchmark algorithms in all environments, especially in low-data regimes. Also notable is that in this strictly batch setting (i.e. where no online sampling whatsoever is permitted), the off-policy adaptations of online algorithms (i.e. DSFN, VDICE) do not perform as consistently as the intrinsically offline ones\textemdash especially DSFN, which involves predicting entire next states (off-policy) for estimating feature maps; this validates some of our original motivations. Finally, note that\textemdash via the joint EBM\textemdash the EDM algorithm readily accommodates (semi-supervised) learning from additional state-only data (with unobserved actions); additional result in Appendix \ref{app:more_experiments}.

\begin{table}[t]\small
\centering
\setlength\tabcolsep{7.35pt}
\renewcommand{\arraystretch}{0.92}
\begin{adjustbox}{max width=\textwidth}
\input{table/results_mimic}
\end{adjustbox}
\vspace{0.5em}
\caption{\textit{Performance Comparison for} \texttt{MIMIC-III}. Action-matching is used to assess the quality of clinical policies learned in both the 2-action and 4-action settings. We report the accuracy of action selection (\textsc{acc}), the area under the receiving operator characteristic curve (\textsc{auc}), and the area under the precision-recall curve (\textsc{apr}).}
\vspace{-2.35em}
\label{tab:mimic}
\end{table}

\vspace{-0.5em}
\section{Discussion}
\vspace{-0.3em}

\dayum{
In this work, we motivated and presented EDM for strictly batch imitation, which retains the simplicity of direct policy learning while accounting for information in visitation distributions. However, we are sampling from an offline model (leveraging multitask learning) of state visitations, not from actual online rollouts (leveraging temporal consistency), so they can only be so useful. The objective also relies on the assumption that samples in $\mathcal{D}$ are sufficiently representative of $\rho_{D}$; while this is standard in literature \cite{kostrikov2019discriminator}, it nonetheless bears reiteration. Our method is agnostic as to discrete/continuous state spaces, but the use of joint EBMs means we only consider categorical actions in this work. That said, the application of EBMs to regression is increasingly of focus \cite{gustafsson2020train}, and future work may investigate the possibility of extending EDM to continuous actions. Overall, our work is enabled by recent advances in joint EBMs, and similarly use contrastive divergence to approximate the KL gradient. Note that EBMs in general may not be the easiest to train, or to gauge learning progress for \cite{grathwohl2020your}. However, for the types of environments we consider, we did not find stability-related issues to be nearly as noticeable as typical of the higher-dimensional imaging tasks EBMs are commonly used for.}



\clearpage

\section*{Broader Impact}

In general, any method for imitation learning has the potential to mitigate problems pertaining to scarcity of expert knowledge and computational resources. For instance, consider a healthcare institution strapped for time and personnel attention\textemdash such as one under the strain of an influx of ICU patients. If implemented as a system for clinical decision support and early warnings, even the most bare-bones policy trained on optimal treatment/monitoring actions has huge potential for streamlining medical decisions, and for allocating attention where real-time clinical judgment is most required.

By focusing our work on the strictly batch setting for learning, we specifically accommodate situations that disallow directly experimenting on the environment during the learning process. This consideration is critical in many conceivable applications: In practice, humans are often on the receiving end of actions and polices, and an imitator policy that must learn by interactive experimentation would be severely hampered due to considerations of cost, danger, or moral hazard. While\textemdash in line with literature\textemdash we illustrate the technical merits of our proposed method with respect to standard control environments, we do take care to highlight the broader applicability of our approach to healthcare settings, as it likewise applies\textemdash without saying\textemdash to education, insurance, or even law enforcement.

Of course, an important caveat is that any method for imitation learning naturally runs the risk of internalizing any existing human biases that may be implicit in the demonstrations collected as training input. That said, a growing field in reinforcement learning is dedicated to maximizing interpretability in learned policies, and\textemdash in the interest of accountability and transparency\textemdash striking an appropriate balance with performance concerns will be an interesting direction of future research.

\section*{Acknowledgments }

We would like to thank the reviewers for their generous and invaluable comments and suggestions. This work was supported by Alzheimer’s Research UK (ARUK), The Alan Turing Institute (ATI) under the EPSRC grant EP/N510129/1, The US Office of Naval Research (ONR), and the National Science Foundation (NSF) under grant numbers 1407712, 1462245, 1524417, 1533983, and 1722516.

\bibliographystyle{unsrt}

\bibliography{neurips_2020}

\appendix
\clearpage

\section{Proofs of Propositions}\label{app:proofs}

\begin{relemma}[restate=interchange,name=]\upshape\label{thm:interchange}
Let $\theta\in\Theta$ be some parameter, consider a random variable $s\in\mathcal{S}$, and fix $f:\mathcal{S}\times\Theta\rightarrow\mathbb{R}$, where $f(s,\theta)$ is continuously differentiable with respect to $\theta$ and integrable for all $\theta$. Assume for some random variable $X$ with finite mean that \smash{$|\tfrac{\partial}{\partial\theta}f(s,\theta)|\leq X$} holds almost surely for all $\theta$. Then:
\begin{equation}
\tfrac{\partial}{\partial\theta}\mathbb{E}[f(s,\theta)]
=
\mathbb{E}[\tfrac{\partial}{\partial\theta}f(s,\theta)]
\end{equation}
\end{relemma}

\textit{Proof}. \smash{$
\tfrac{\partial}{\partial\theta}\mathbb{E}[f(s,\theta)]
=
\lim_{\delta\rightarrow0}\tfrac{1}{\delta}(\mathbb{E}[f(s,\theta+\delta)]-\mathbb{E}[f(s,\theta)])
=
\lim_{\delta\rightarrow0}\mathbb{E}[\tfrac{1}{\delta}(f(s,\theta+\delta)$$-$$f(s,\theta))]
$}
\smash{$
=
\lim_{\delta\rightarrow0}\mathbb{E}[\tfrac{\partial}{\partial\theta}f(s,\tau(\delta))]
=
\mathbb{E}[\lim_{\delta\rightarrow0}\tfrac{\partial}{\partial\theta}f(s,\tau(\delta))]
=
\mathbb{E}[\tfrac{\partial}{\partial\theta}f(s,\theta)]
$}, where for the third equality the mean value theorem guarantees the existence of $\tau(\delta)\in(\theta,\theta+\delta)$, and the fourth equality uses the dominated convergence theorem where \smash{$|\tfrac{\partial}{\partial\theta}f(s,\tau(\delta))|\leq X$} by assumption. Note that generalizing to the multivariate case (i.e. gradients) simply requires that the bound be on \smash{$\max_{i}|\tfrac{\partial}{\partial\theta_{i}}f(s,\theta)|$} for elements $i$ of $\theta$. Note that most machine learning models (and energy-based models) meet/assume these regularity conditions or similar variants; see e.g. discussion presented in Section 18.1 in \cite{goodfellow2016deep}.

\surrogate*

\textit{Proof}. For each $s$, first write the state occupancy measure as \smash{$\rho_{\theta}(s)=e^{-E_{\theta}(s)}/\int_{\mathcal{S}}e^{-E_{\theta}(s)}ds$}, so:
\begin{equation}
-\log\rho_{\theta}(s)=E_{\theta}(s)+\log\textstyle\int_{\mathcal{S}}e^{-E_{\theta}(s)}ds
\end{equation}

\vspace{-1.3em}

with gradients given by:
\begin{align}
-\nabla_{\theta}\log\rho_{\theta}(s)&=\nabla_{\theta}E_{\theta}(s)+\nabla_{\theta}\log\textstyle\int_{\mathcal{S}}e^{-E_{\theta}(s)}ds \nonumber \\
&=\nabla_{\theta}E_{\theta}(s)-\frac{\int_{\mathcal{S}}\nabla_{\theta}E_{\theta}(s)e^{-E_{\theta}(s)}ds}{\int_{\mathcal{S}}e^{-E_{\theta}(s)}ds} \nonumber \\
&=\nabla_{\theta}E_{\theta}(s)-\mathbb{E}_{s\sim\rho_{\theta}}\nabla_{\theta} E_{\theta}(s)
\end{align}
Then taking expectations over $\rho_{D}$ and substituting in the energy term per Equation \ref{eqn:energy}, we have that:
\begin{align}
-\nabla_{\theta}\mathbb{E}_{s\sim\rho_{D}}\log\rho_{\theta}(s)&=\mathbb{E}_{s\sim\rho_{D}}\big[\nabla_{\theta}E_{\theta}(s)-\mathbb{E}_{s\sim\rho_{\theta}}\nabla_{\theta}E_{\theta}(s)\big] \nonumber \\
&=\mathbb{E}_{s\sim\rho_{D}}\nabla_{\theta}E_{\theta}(s)-\mathbb{E}_{s\sim\rho_{\theta}}\nabla_{\theta}E_{\theta}(s) \nonumber \\
&=\mathbb{E}_{s\sim\rho_{\theta}}\nabla_{\theta}\big(\log\textstyle\sum_{a}e^{f_{\theta}(s)[a]}\big)-\mathbb{E}_{s\sim\rho_{D}}\nabla_{\theta}\big(\log\textstyle\sum_{a}e^{f_{\theta}(s)[a]}\big) \nonumber \\
&=\nabla_{\theta}\big(\mathbb{E}_{s\sim\rho_{\theta}}\log\textstyle\sum_{a}e^{f_{\theta}(s)[a]}-\mathbb{E}_{s\sim\rho_{D}}\log\textstyle\sum_{a}e^{f_{\theta}(s)[a]}\big) \nonumber \\
&=\nabla_{\theta}\mathcal{L}_{\rho}(\theta)
\end{align}
where the fourth equality uses Lemma \ref{thm:interchange}. Hence we can define \smash{$\mathcal{L}_{\rho}(\theta)\doteq\mathbb{E}_{s\sim\rho_{D}}E_{\theta}(s)-\mathbb{E}_{s\sim\rho_{\theta}}E_{\theta}(s)$} in lieu of the first term in Equation \ref{eqn:obj_loss_original}. However, note that the (gradient-based) implementation of Algorithm 1 works even without first obtaining an expression for $\mathcal{L}_{\rho}(\theta)$ per se, and is correct due to a simpler reason: The batched (empirical loss) \smash{$\nabla_{\theta}\hat{\mathcal{L}}_{\rho}$} portion of the update (Line 9) is directly analogous to the gradient update in standard contrastive divergence; see e.g. Section 18.2 in \cite{goodfellow2016deep}.
\QED

Propositions \ref{thm:classical}--\ref{thm:bcedm} first require an additional lemma that allows moving freely between the space of (soft) $Q$-functions and reward functions. Recall the (soft) Bellman operator $\mathbb{B}^{*}_{R}:\mathbb{R}^{\mathcal{S}\times\mathcal{A}}$\pix$\rightarrow\mathbb{R}^{\mathcal{S}\times\mathcal{A}}$:
\begin{equation}
(\mathbb{B}^{*}_{R}Q)(s,a)=R(s,a)+\gamma\mathbb{E}_{T}[\text{softmax}_{a^{\prime}}Q(s^{\prime},a^{\prime})|s,a]
\end{equation}
where $\text{softmax}_{a}Q(s,a)\doteq\log\sum_{a}e^{Q(s,a)}$. We know that $\mathbb{B}^{*}_{R}$ is contractive with $Q^{*}_{R}$ its unique fixed point \cite{ziebart2010modeling, haarnoja2017reinforcement}. Now, let us define the (soft) inverse Bellman operator $\mathbb{J}^{*}:\mathbb{R}^{\mathcal{S}\times\mathcal{A}}\rightarrow\mathbb{R}^{\mathcal{S}\times\mathcal{A}}$ such that:
\begin{equation}
(\mathbb{J}^{*}Q)(s,a)=Q(s,a)-\gamma\mathbb{E}_{T}[\text{softmax}_{a}Q(s^{\prime},a^{\prime})|s,a]
\end{equation}

\begin{relemma}[restate=biject,name=]\upshape\label{thm:biject}
The operator $\mathbb{J}^{*}$ is \textit{bijective}: $Q=Q^{*}_{R}\Leftrightarrow\mathbb{J}^{*}Q=R$, hence we can write $(\mathbb{J}^{*})^{-1}R=Q^{*}_{R}$. This is the ``soft'' version of an analogous statement made for ``hard'' optimality first shown in \cite{piot2017bridging}.
\end{relemma}

\textit{Proof}. By the uniqueness of the fixed point of $\mathbb{B}^{*}_{R}$, we have that $R=\mathbb{J}^{*}Q\Leftrightarrow \mathbb{B}^{*}_{R}Q=Q\Leftrightarrow Q=Q^{*}_{R}$. Therefore the inverse image of every singleton $R\in\mathbb{R}^{\mathcal{S}\times\mathcal{A}}$ must exist, and is uniquely equal to $Q^{*}_{R}$. This argument is the direct counterpart to Theorem 2 in \cite{piot2017bridging}\textemdash which uses argmax instead of softmax.

\classical*

\textit{Proof}. From Equations \ref{eqn:obj_il} and \ref{eqn:policy_class}, choosing $\mathcal{L}$ to be the forward KL divergence yields the following:
\begin{equation}
\textstyle\argmax_{\theta}\big(\mathbb{E}_{s\sim\rho_{\theta}}\mathbb{E}_{a\sim\pi_{D}(\cdot|s)}f_{\theta}(s)[a]-\mathbb{E}_{s\sim\rho_{\theta}}\log\sum_{a}e^{f_{\theta}(s)[a]}\big)
\end{equation}
Now, observe that we are free to identify the logits $f_{\theta}(s)[a]\in\mathbb{R}^{\mathcal{S}\times\mathcal{A}}$ with a (soft) $Q$-function. Specifically, define $Q(s,a)\doteq f_{\theta}(s)[a]$ for all $s,a\in\mathcal{S}\times\mathcal{A}$. Then by Lemma \ref{thm:biject} we know there exists a unique $R\in\mathbb{R}^{\mathcal{S}\times\mathcal{A}}$ that $\mathbb{J}^{*}$ takes $Q$ to. Hence $f_{\theta}(s)[a]=Q^{*}_{R}(s,a)$ for some $R$, and we can write:
\begin{equation}
\textstyle\argmax_{R}\big(\mathbb{E}_{s\sim\rho^{*}_{R}}\mathbb{E}_{a\sim\pi_{D}(\cdot|s)}Q^{*}_{R}(s,a)-\mathbb{E}_{s\sim\rho^{*}_{R}}\log\sum_{a}e^{Q^{*}_{R}(s,a)}\big)
\end{equation}
where $\pi^{*}_{R}(a|s)=e^{Q^{*}_{R}(s,a)-V^{*}_{R}(s)}$. Then Proposition \ref{thm:classical} follows, since $V^{*}_{R}(s)=\log\sum_{a}e^{Q^{*}_{R}(s,a)}$.
\QED

\bcedm*

\textit{Proof}. By definition of behavioral cloning, the only difference is that the expectation in Equation \ref{eqn:obj_il} is taken over $\rho_{D}$; then the same argument for Proposition \ref{thm:classical} applies. As for EDM, from Equation \ref{eqn:obj_loss_surrogate}:
\begin{align}
\mathcal{L}_{\text{surr}}(\theta)&=\mathcal{L}_{\rho}(\theta)+\mathcal{L}_{\pi}(\theta) \nonumber \\
&=\mathbb{E}_{s\sim\rho_{D}}E_{\theta}(s)-\mathbb{E}_{s\sim\rho_{\theta}}E_{\theta}(s)
-
\mathbb{E}_{s,a\sim\rho_{D}}\log\pi_{\theta}(a|s)
\nonumber \\
&=\mathbb{E}_{s\sim\rho_{\theta}}\log\textstyle\sum_{a}e^{f_{\theta}(s)[a]}-\mathbb{E}_{s\sim\rho_{D}}\log\textstyle\sum_{a}e^{f_{\theta}(s)[a]} \nonumber \\
&~~~~+\mathbb{E}_{s,a\sim\rho_{D}}\log\textstyle\sum_{a}e^{f_{\theta}(s)[a]}-\mathbb{E}_{s,a\sim\rho_{D}}f_{\theta}(s)[a] \nonumber \\
&=\mathbb{E}_{s\sim\rho_{\theta}}\log\textstyle\sum_{a}e^{f_{\theta}(s)[a]}-\mathbb{E}_{s,a\sim\rho_{D}}f_{\theta}(s)[a]
\end{align}
Therefore minimizing $\mathcal{L}_{\text{surr}}(\theta)$ is equivalent to:
\begin{equation}
\textstyle\argmax_{\theta}\big(\mathbb{E}_{s\sim\rho_{D}}\mathbb{E}_{a\sim\pi_{D}(\cdot|s)}f_{\theta}(s)[a]-\mathbb{E}_{s\sim\rho_{\theta}}\log\textstyle\sum_{a}e^{f_{\theta}(s)[a]}\big)
\end{equation}
From this point onwards, the same strategy for Proposition \ref{thm:classical} again applies, completing the proof.
\QED

\section{Experiment Details}\label{app:more_experiments}

\textbf{Gym Environments}~
Environments used for experiments are from OpenAI gym \cite{brockman2016openai}. Table \ref{tab:envs} shows environment names and version numbers, dimensions of each observation space, and cardinalities of each action space. Each environment is associated with a true reward function (unknown to all imitation algorithms). In each case, the ``expert'' demonstrator is obtained using a pre-trained and hyperparameter-optimized agent from the RL Baselines Zoo \cite{raffin2018rl} in Stable OpenAI Baselines \cite{hill2018stable}; for all environments, demonstration datasets $\mathcal{D}$ are generated using the PPO2 agent \cite{schulman2017proximal} trained on the true reward function, with the exception of \texttt{CartPole}, for which we use the DQN agent (which we find performs better than PPO2). Performance of demonstrator and random policies are shown:

\begin{table}[h]\small
\newcolumntype{H}{>{          \arraybackslash}m{2.4 cm}}
\newcolumntype{I}{>{\centering\arraybackslash}m{2.8 cm}}
\newcolumntype{J}{>{\centering\arraybackslash}m{2.0 cm}}
\newcolumntype{K}{>{\centering\arraybackslash}m{2.0 cm}}
\newcolumntype{L}{>{\centering\arraybackslash}m{2.3 cm}}
\newcolumntype{M}{>{\centering\arraybackslash}m{2.4 cm}}
\setlength\tabcolsep{1.35pt}
\renewcommand{\arraystretch}{1.02}
\begin{adjustbox}{max width=\textwidth}
\input{table/environments}
\end{adjustbox}
\vspace{0.5em}
\caption{\textit{Details of Environments}. Demonstrator and random performances are computed using 1,000 episodes.}
\vspace{-2.0em}
\label{tab:envs}
\end{table}

\textbf{Healthcare Environments}~
\texttt{MIMIC-III} is a real-world medical dataset consisting of patients treated in intensive care units from the \href{https://mimic.physionet.org}{Medical Information Mart for Intensive Care} \cite{johnson2016mimic}, which records physiological data streams for over 22,000 patients. We extract the records for ICU patients administered with antibiotic treatment and/or mechanical ventilation (5,833 in total). For each patient, we define the observation space to be the 28 most frequently measured patient covariates from the past two days, including vital signs (e.g. temperature, heart rate, blood pressure, oxygen saturation, respiratory rate, etc.) and lab tests (e.g. white blood cell count, glucose levels, etc.), aggregated on a daily basis during their ICU stay. Each patient trajectory has up to 20 time steps. In this environment, the action space consists of the possible treatment choices administered by the doctor every day over the course of the patient's ICU stay, and the ``expert'' demonstrations are simply the trajectories of states and actions recorded in the dataset. We consider two versions of \texttt{MIMIC-III}; one with 2 actions: with ventilator support, or no treatment (\texttt{MIMIC-III-2a}), and another with 4 actions: with ventilator support, antibiotics treatment, ventilator support plus antibiotics, or no treatment (\texttt{MIMIC-III-4a}).

\textbf{Detailed Results}~
Exact experiment results are shown in Table \ref{tab:main}. For each combination of gym environment, imitation algorithm, and dataset size, we follow convention for randomization in our experiment setup by rolling out multiple trajectories ($n_{\text{traj}}$) per trained policy, seeding the experiment multiple times with different expert demonstrations ($n_{\text{demo}}$), and training multiple such policies from different random initializations ($n_{\text{init}}$); see e.g. \cite{ho2016generative}. Here we set $n_{\text{traj}}$$=$$300$, $n_{\text{demo}}$$=$$10$, and $n_{\text{init}}$$=$$5$. Table \ref{tab:main} shows the means of performance metrics, as well as their standard errors; for ease of comparison, all numbers for gym environments are scaled (according to the performance of demonstrator and random policies given in Table \ref{tab:envs}) such that the demonstrator attains a return of 1 and the random policy attains a return of 0. For the real-world healthcare environments, we have no access to the ground-truth reward function, and we cannot perform live policy rollouts. We therefore assess imitation performance according to action-matching on held-out test trajectories; see e.g. \cite{lee2019github}. In each of $n_{\text{demo}}$$=$$10$ folds, we use an 80\%-20\% train-test split (i.e. 4,666 patients for training, and 1,167 held out for testing). In each instance, we report accuracy of action selection (\textsc{acc}), area under the receiving operator characteristic curve (\textsc{auc}), and area under the precision-recall curve (\textsc{apr}).

\begin{table}[h]\small
\centering
\setlength\tabcolsep{7.35pt}
\renewcommand{\arraystretch}{0.99}
\begin{adjustbox}{max width=\textwidth}
\input{table/results_main}
\end{adjustbox}
\vspace{0.5em}
\caption{\textit{Detailed Results for Gym and Healthcare Environments}. Bold numbering indicates best performance.}
\vspace{-2.0em}
\label{tab:main}
\end{table}

\textbf{Implementations}~
Wherever possible, policies trained by all imitation algorithms share the same policy network architecture: two hidden (fully connected) layers of 64 units each, followed by ELU activations, or\textemdash for Atari\textemdash a convolutional neural network with 3 (convolutional) layers of 32-64-64 filters, followed by a fully connected layer with 64 units, with all layers followed by ReLU activations. For all environments, we use the Adam optimizer with batch size 64, 10k iterations, and learning rate 1e-3. Except explicitly standardizing policy networks across imitation algorithms, all comparators are implemented via the original publicly available source code. Where applicable, we use the optimal hyperparameters in the original implementations. The source code for EDM is found at {\small\url{https://bitbucket.org/mvdschaar/mlforhealthlabpub}} \& {\small\url{https://github.com/danieljarrett/EDM}}.

\textbf{Hyperparameters for EDM}~
Algorithm \ref{alg:edm} is implemented using the source code for joint EBMs \cite{grathwohl2020your} publicly available at \href{https://github.com/wgrathwohl/JEM}{https://github.com/wgrathwohl/JEM}. Instead of Wide-Resnet, for \texttt{Acrobot}, \texttt{Cartpole}, \texttt{LunarLander}, \texttt{MIMIC-III-2a}, and \texttt{MIMIC-III-4a} we use the fully-connected policy network above, and for \texttt{BeamRider} the convolutional neural network above. Specific to EDM are the joint EBM training hyperparameters, which we inherit from \cite{grathwohl2020your, grathwohl2020github}: noise coefficient $\sigma$$=$$0.01$, buffer size $\kappa$$=$$10000$, length $\iota$$=$$20$, and reinitialization $\delta$$=$$0.05$. We find that these default settings work well with SGLD step size $\alpha$$=$$0.01$; for further EBM training-related discussions, we refer to \cite{grathwohl2020your, du2019implicit}.

\textbf{Hyperparameters for VDICE}~
We take the original source code of \cite{kostrikov2020imitation}, which is publicly available at \href{https://github.com/google-research/google-research/tree/master/value\_dice}{https://github.com/google-research/google-research/tree/master/value\_dice}. In order to adapt the model to work with discrete action spaces, we use a Gumbel-softmax parameterization for the last layer of the actor network. For \texttt{Acrobot}, \texttt{Cartpole}, \texttt{LunarLander}, \texttt{MIMIC-III-2a}, and \texttt{MIMIC-III-4a} both the actor architecture and the discriminator architecture has two hidden (fully connected) layers of 64 units each with ReLU activation, and\textemdash for Atari\textemdash the actor and discriminator are replaced with convolutional neural networks with 3 (convolutional) layers of 32-64-64 filters followed by a fully connected layer with 64 units, with all layers followed by ReLU activations. Per the original design, the output is concatenated with the action; this is then passed through 2 additional hidden layers with 64 units each. In addition, to enable strictly batch learning, we set the ``replay regularization'' coefficient to zero. Furthermore, the actor network is regularized with an ``orthogonal regularization'' coefficient of 1e-4, actor learning rate of 1e-5, and discriminator learning rate of 1e-3.

\textbf{Hyperparameters for DSFN}~
We take the original source code of \cite{kostrikov2020imitation}, which is publicly available at \href{https://github.com/dtak/batch-apprenticeship-learning}{https://github.com/dtak/batch-apprenticeship-learning}. Per \cite{lee2019truly}, for \texttt{Acrobot}, \texttt{Cartpole}, \texttt{LunarLander}, \texttt{MIMIC-III-2a}, and \texttt{MIMIC-III-4a} we use a ``warm-start'' policy network with two shared layers of 128 and 64 dimensions and tanh activation. The hidden layer of size 64 is used as the feature map in the IRL algorithm. Each multitask head in the warm-start policy network has a hidden layer with 128 units and tanh activation. The DQN network (i.e. for learning the optimal policy given a set of reward weights) has 2 hidden (fully-connected) layers with 64 units each, and likewise the DSFN network for estimating feature expectations also has 2 hidden (fully-connected) layers with 64 units. For \texttt{BeamRider}, the first hidden layer in the warm-start policy network is replaced by a convolutional neural network with 3 layers of 32-64-64 filters, and the DQN and DSFN networks are also replaced by the convolutional neural network above. For all environments, the warm-start policy network is trained for 50k steps with the Adam optimizer, learning rate 3e-4, and batch size 64. The DQN network is trained for 30k steps with learning rate 3e-4 and batch size 64 (Adam). Finally, the DSFN network is trained for 50,000 iterations with the learning rate 3e-4 and batch size 32 (Adam).

\textbf{Hyperparameters for RCAL}~
This augments the policy loss with an additional sparsity-based loss on the implied rewards \smash{$\hat{R}(s,a)\doteq f_{\theta}(s)[a]-\gamma\text{softmax}_{a^{\prime}}f_{\theta}(s^{\prime})[a^{\prime}]$} obtained by inverting the Bellman equation \cite{piot2014boosted, piot2017bridging}. For \texttt{Acrobot}, \texttt{Cartpole}, \texttt{LunarLander}, \texttt{MIMIC-III-2a}, and \texttt{MIMIC-III-4a} we use the fully-connected policy network described above, and for \texttt{BeamRider} the convolutional neural network above. Specific to RCAL is its sparsity-based regularization coefficient, which is set at 1e-2.

\textbf{Hyperparameters for BC}~
The only difference between BC and EDM is the presence of $\mathcal{L}_{\rho}$, which we remove for our implementation of BC. (Unlike e.g. \cite{piot2017bridging}, we do not consider more primitive methods such as linear classifiers/trees to serve as BC, which would not make for a fair comparison/ ablation). For \texttt{Acrobot}, \texttt{Cartpole}, \texttt{LunarLander}, \texttt{MIMIC-III-2a}, and \texttt{MIMIC-III-4a} we use the fully-connected policy network above, and for \texttt{BeamRider} the convolutional neural network above.

\begin{wrapfigure}{t}{0.33\textwidth}\small
\centering
\vspace{-1.0em}
\vspace{-0.15em}
\adjustbox{trim={0.02\width} {0.05\height} {0.05\width} {0.02\height},clip}{
\includegraphics[width=\linewidth, trim=8em 0em 7em 15em]{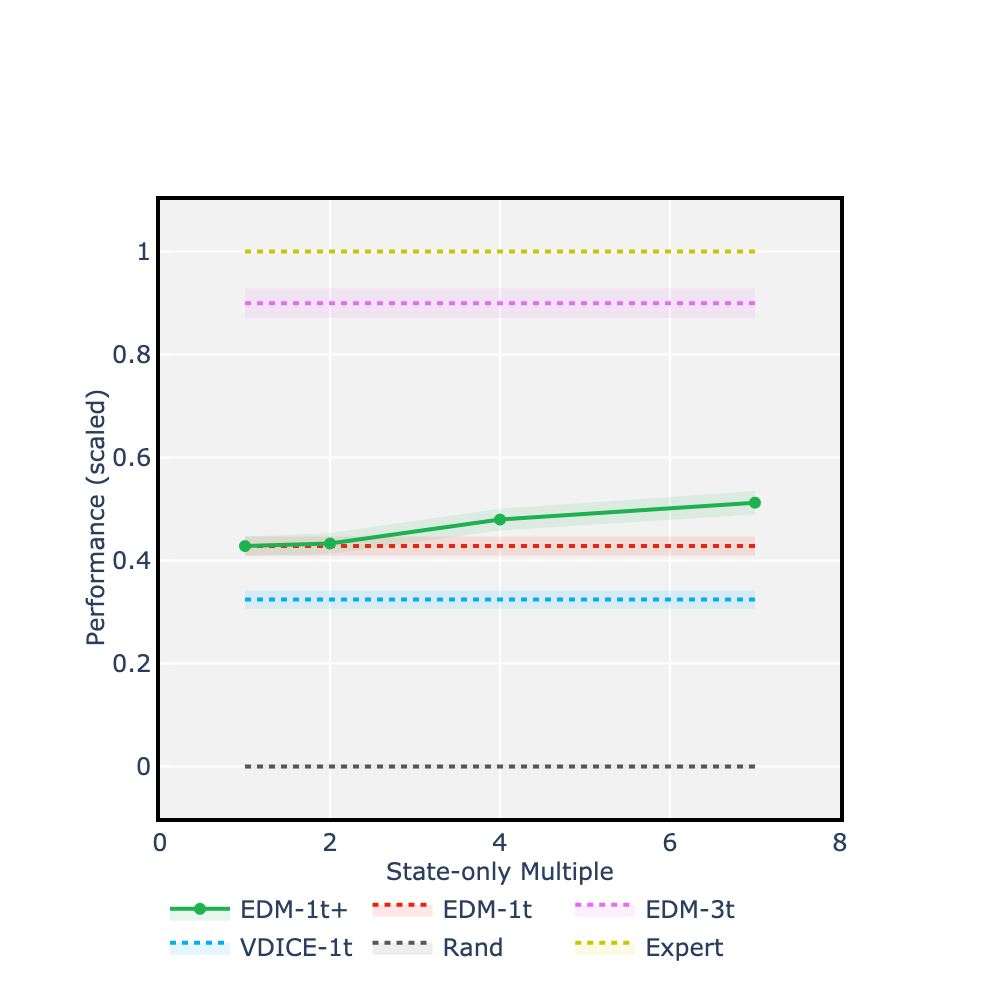}}
\caption{\textit{Semi-Supervised Learning}.}
\vspace{-0.5em}
\label{fig:semi}
\end{wrapfigure}

\textbf{Semi-Supervised Learning}~
While this is beyond the scope of this work, we briefly note that\textemdash by analogy to joint energy-based modeling in general\cite{grathwohl2020your}\textemdash the EDM algorithm can additionally benefit from semi-supervised learning. Specifically, consider a data-scarce setting where we only have access to limited state-action pairs from the demonstrator\textemdash but may have access to additional state-only data. Broadly, this situation arises whenever states are more conveniently observed than actions are. For \texttt{CartPole}, Figure \ref{fig:semi} shows the results of the original EDM trained on one demonstrator trajectory's worth of state-action pairs, but with access to additional state-only data (\textbf{EDM-1t+}) shown in the $x$-axis as multiples of the original amount of state-action data. For comparison, we also reference the performance of EDM without such additional state-only data (EDM-1t), as well as the performance of its closest competitor (VDICE-1t), both trained on one trajectory's worth of state-action pairs alone. Notably, observe that (purely by dint of state-only distribution matching) EDM-1t+ manages to extract a sizable gain in performance as the amount of state-only data available increases up to seven-fold. While this improvement is\textemdash as expected\textemdash less than that conferred by simply adding more state-action trajectories (cf. EDM-3t, which is trained on 3 trajectories' worth of state-action pairs), simply adding state-only data manages to provide as much of a performance boost as the original difference between EDM and VDICE (trained on one trajectory's worth of state-action pairs).

\vspace{-0.5em}
\section{Further Related Work}\label{app:more_related}
\vspace{-0.3em}

Throughout this work, we discussed the goal of imitation learning \cite{le2016smooth, hussein2017imitation, yue2018imitation} in the strictly batch setting, behavioral cloning \cite{pomerleau1991efficient, bain1999framework, syed2010reduction, ross2010efficient} and its relatives \cite{syed2007imitation, ross2011reduction, piot2014boosted, piot2017bridging, attia2018global}, and relationships with the apprenticeship learning family of techniques, including classic (online) inverse reinforcement learning \cite{ng2000algorithms, abbeel2004apprenticeship, neu2007apprenticeship, ramachandran2007bayesian, syed2008game, ziebart2008maximum, babes2011apprenticeship, choi2011map}, (online) adversarial imitation learning \cite{ho2016generative, baram2017model, jeon2018bayesian, finn2016connection, fu2018learning, qureshi2019adversarial, ghasemipour2019divergence, kim2018imitation, xiao2019wasserstein}, as well as their respective off-policy relatives \cite{klein2011batch, mori2011model, klein2012inverse, tanwani2013inverse, tossou2013probabilistic, herman2016inverse, jain2019model, lee2019truly, blonde2019sample, kostrikov2019discriminator, kostrikov2020imitation}. Table \ref{tab:related} summarizes the major aspects of these works as pertinent to our discussion and development.

Further to these works, we also note that another line of research on (online) imitation learning seeks to incentivize the imitating policy to remain within the distribution/support of states encountered in the expert demonstrations \cite{choi2016density, schroecker2017state, liu2020state, wang2019random, brantley2020disagreement, reddy2020sqil, dadashi2020primal, liu2020energy}. For instance, this is approached through random expert distillation \cite{wang2019random}, through ensembles of agents \cite{brantley2020disagreement}, or the simple and elegant approach of assigning a unit reward to all demonstrated actions that occur in demonstrated states, and zero otherwise \cite{reddy2020sqil}. In general, these methods follow a ``two-step'' formula, where in the first step some notion of a surrogate reward function is derived/defined, and in the second step this reward function is optimized by way of environment interactions (and as such, they are inherently online techniques). In the same vein, while \cite{liu2020energy} bears some superficial resemblance to our method by way of energy-based modeling, it is an inherently online technique that depends on training an agent against an explicitly estimated reward function: In the first step, their reward function is defined by modeling the negative energy of the (joint) state-action distribution. However, as with the aforementioned two-step approaches, this must then be followed by an online optimization of this reward function\textemdash and is therefore inoperable in our strictly batch setting. Moreover, not unlike in adversarial imitation learning, their KL-divergence minimization interpretation similarly requires the assumption that the optimal reward function is indeed attained\textemdash an issue our formulation does not encounter. In contrast, EDM works by decomposing the state-action distribution into an (explicit) policy term and an (implicit) state visitation distribution term, resulting in a single optimization that works in an entirely offline manner.

Finally, tangentially related to our work is a family of inverse reinforcement learning methods designed for reward learning in an offline, model-free setting \cite{pirotta2016inverse, tateo2017gradient, metelli2017compatible}. However, they require access to the demonstrator's policy itself to begin with, and their objective is rather in the inverse problem \textit{per se}\textemdash that is, of explicitly recovering the underlying reward function in order to understand behavior.

{\small
\bibliography{neurips_2020}
}

\end{document}

%% file: table/related_work.tex
\begin{tabular}{O|ABCDEFG}
\toprule
\multicolumn{2}{c}{\textbf{Formulation}} & \textbf{Example}
& {Paramterized \smash{$\hphantom{^{X}}$}Policy\smash{\pix$^{\text{(1)}}$}}
& {Non-Restrictive \smash{$\hphantom{^{X}}$}Regularizer\smash{\pix$^{\text{(1)}}$}}
& {Dynamics \smash{$\hphantom{^{3}}$}Awareness\smash{\pix$^{\text{(2)}}$}}
& {Operable \smash{$\hphantom{^{3}}$}Strictly~Batch\smash{\pix$^{\text{(3)}}$}}
& {Directly \smash{$\hphantom{^{3}}$}Optimized\smash{\pix$^{\text{(3)}}$}} \\
\midrule
\parbox[t]{2mm}{\multirow{3}{*}{\pix\rotatebox[origin=c]{90}{Online (Original)\pix}}}
& ~~Max Margin            & \cite{ng2000algorithms, abbeel2004apprenticeship}      & \xm & \xm & \cm & \xm & \xm \\
& ~~Minimax Game          & \cite{syed2008game}                                    & \xm & \xm & \cm & \xm & \xm \\
& ~~Min Policy Loss       & \cite{neu2007apprenticeship}                           & \xm & \xm & \cm & \xm & \xm \\
& ~~Max Likelihood        & \cite{babes2011apprenticeship}                         & \xm & \xm & \cm & \xm & \xm \\
& ~~Max Entropy           & \cite{ziebart2008maximum, ziebart2010modeling}         & \xm & \xm & \cm & \xm & \xm \\
& ~~Max A Posteriori      & \cite{ramachandran2007bayesian, choi2011map}           & \xm & \xm & \cm & \xm & \xm \\
& ~~Adversarial Imitation & \cite{ho2016generative, baram2017model, jeon2018bayesian, finn2016connection, fu2018learning, qureshi2019adversarial, ghasemipour2019divergence}
                                                                                   & \cm & \cm & \cm & \xm & \xm \\
\midrule
\parbox[t]{2mm}{\multirow{3}{*}{\pix\rotatebox[origin=c]{90}{Off. (Adaptation)}}}
& ~~Max Margin            & \cite{lee2019truly, klein2011batch}                    & \xm & \xm & \cm & \cm & \xm \\
& ~~Minimax Game          & \cite{mori2011model}                                   & \xm & \xm & \cm & \cm & \xm \\
& ~~Min Policy Loss       & \cite{klein2012inverse}                                & \xm & \xm & \cm & \cm & \cm \\
& ~~Max Likelihood        & \cite{jain2019model}                                   & \xm & \xm & \cm & \cm & \xm \\
& ~~Max Entropy           & \cite{herman2016inverse}                               & \xm & \xm & \cm & \cm & \xm \\
& ~~Max A Posteriori      & \cite{tossou2013probabilistic}                         & \xm & \xm & \cm & \cm & \cm \\
& ~~Adversarial Imitation & \cite{kostrikov2020imitation}                          & \cm & \cm & \cm & \cm & \xm \\
\midrule
\multicolumn{2}{l}{Behavioral Cloning}    & \cite{ross2010efficient} & \cm & \xm & \xm & \cm & \cm \\
\multicolumn{2}{l}{Reward-Regularized BC} & \cite{piot2014boosted}   & \cm & \xm & \cm & \cm & \cm \\
\midrule
\multicolumn{2}{l}{\textbf{EDM}}          & \textbf{(Ours)}          & \cm & \cm & \cm & \cm & \cm \\
\bottomrule
\end{tabular}

%% file: table/results_mimic.tex
\begin{tabular}{l|ccc|ccc}
\toprule
& \multicolumn{3}{c|}{\textbf{2-Action Setting} (Ventilator Only)} &
\multicolumn{3}{c}{\textbf{4-Action Setting} (Antibiotics + Vent.)} \\
\midrule
\textit{Metrics} & {ACC} & {AUC} & {APR} & {ACC} & {AUC} & {APR} \\
\midrule
BC    & 0.861\pmm0.013 & 0.914\pmm0.003 & 0.902\pmm0.005 & 0.696\pmm0.006 & 0.859\pmm0.003 & 0.659\pmm0.007 \\
RCAL  & 0.872\pmm0.007 & 0.911\pmm0.007 & 0.898\pmm0.006 & 0.701\pmm0.007 & 0.864\pmm0.003 & 0.667\pmm0.006 \\
DSFN  & 0.865\pmm0.007 & 0.906\pmm0.003 & 0.885\pmm0.001 & 0.682\pmm0.005 & 0.857\pmm0.002 & 0.665\pmm0.003 \\
VDICE & 0.875\pmm0.004 & 0.915\pmm0.001 & 0.904\pmm0.002 & 0.707\pmm0.005 & 0.864\pmm0.002 & 0.673\pmm0.003 \\
Rand  & 0.498\pmm0.007 & 0.500\pmm0.000 & 0.500\pmm0.000 & 0.251\pmm0.005 & 0.500\pmm0.000 & 0.250\pmm0.000 \\
\midrule
\textbf{EDM}& \textbf{0.891\pmm0.004} & \textbf{0.922\pmm0.004} & \textbf{0.912\pmm0.005} & \textbf{0.720\pmm0.007} & \textbf{0.873\pmm0.002} & \textbf{0.681\pmm0.008} \\
\bottomrule
\end{tabular}

%% file: table/environments.tex
\begin{tabular}{H|IJKLM}
\toprule
\textit{Environments} & {Observation Space} & {Action Space} & {Demonstrator} & {Random Perf.} & {Demonstrator Perf.} \\
\midrule
\texttt{CartPole-v1}    & Continuous (4)                       & Discrete (2) & DQN Agent    & 19.12\pmm1.76    & 500.00\pmm0.00    \\
\texttt{Acrobot-v1}     & Continuous (6)                       & Discrete (3) & PPO2 Agent   & -439.92\pmm13.14 & -87.32\pmm12.02   \\
\texttt{LunarLander-v2} & Continuous (8)                       & Discrete (4) & PPO2 Agent   & -452.22\pmm61.24 & 271.71\pmm17.88   \\
\texttt{BeamRider-v4}   & ~Cont.\pix~(210$\times$160$\times$3) & Discrete (9) & PPO2 Agent  & 954.84\pmm214.85 & 1623.80\pmm482.27 \\
\texttt{MIMIC-III-2a}   & ~Continuous (56)                     & Discrete (2) & Human Agent & -                & -                 \\
\texttt{MIMIC-III-4a}   & ~Continuous (56)                     & Discrete (4) & Human Agent & -                & -                 \\
\bottomrule
\end{tabular}

%% file: table/results_main.tex
\begin{tabular}{l|c|ccccc}
\toprule
\multicolumn{2}{l|}{} & {BC} & {RCAL} & {DSFN} & {VDICE} & {EDM} \\
\midrule
{} & {\textit{Demos}} & \multicolumn{5}{c}{\textit{Average Returns}} \\
\midrule
\pix\texttt{Acrobot-v1}
 &  1 & 0.796\pmm0.078 & 0.422\pmm0.082 & 0.062\pmm0.141 & 0.857\pmm0.045 & \textbf{0.896\pmm0.064} \\
 &  3 & 0.976\pmm0.028 & 0.832\pmm0.066 & 0.227\pmm0.128 & 0.947\pmm0.033 & \textbf{0.998\pmm0.026} \\
 &  7 & 0.981\pmm0.028 & 0.975\pmm0.034 & 0.489\pmm0.075 & 0.953\pmm0.036 & \textbf{0.999\pmm0.026} \\
 & 10 & 0.986\pmm0.029 & 0.990\pmm0.030 & 0.601\pmm0.076 & 0.967\pmm0.032 & \textbf{0.999\pmm0.025} \\
 & 15 & 0.994\pmm0.028 & 0.997\pmm0.028 & 0.825\pmm0.050 & 0.976\pmm0.031 & \textbf{1.000\pmm0.026} \\
\midrule
\pix\texttt{CartPole-v1}
 &  1 & 0.321\pmm0.026 & 0.233\pmm0.036 & 0.317\pmm0.013 & 0.324\pmm0.018 & \textbf{0.428\pmm0.019} \\
 &  3 & 0.607\pmm0.048 & 0.586\pmm0.043 & 0.373\pmm0.073 & 0.738\pmm0.028 & \textbf{0.900\pmm0.029} \\
 &  7 & 0.819\pmm0.041 & 0.894\pmm0.027 & 0.523\pmm0.081 & 0.867\pmm0.022 & \textbf{0.982\pmm0.011} \\
 & 10 & 0.932\pmm0.026 & 0.991\pmm0.007 & 0.458\pmm0.047 & 0.967\pmm0.013 & \textbf{1.000\pmm0.001} \\
 & 15 & 0.997\pmm0.003 & \textbf{0.998\pmm0.001} & 0.653\pmm0.074 & 0.995\pmm0.004 & \textbf{0.998\pmm0.002} \\
\midrule
\pix\texttt{LunarLander-v2}
 &  1 & 0.575\pmm0.071 & 0.540\pmm0.090 & 0.229\pmm0.104 & 0.255\pmm0.071 & \textbf{0.633\pmm0.081} \\
 &  3 & 0.869\pmm0.055 & 0.875\pmm0.055 & 0.698\pmm0.050 & 0.385\pmm0.063 & \textbf{0.889\pmm0.069} \\
 &  7 & 0.938\pmm0.035 & 0.914\pmm0.057 & 0.776\pmm0.053 & 0.411\pmm0.063 & \textbf{0.956\pmm0.044} \\
 & 10 & 0.961\pmm0.035 & 0.952\pmm0.047 & 0.887\pmm0.042 & 0.418\pmm0.059 & \textbf{0.966\pmm0.040} \\
 & 15 & 0.968\pmm0.028 & \textbf{0.970\pmm0.028} & 0.913\pmm0.032 & 0.417\pmm0.054 & \textbf{0.970\pmm0.033} \\
\midrule
\pix\texttt{BeamRider-v4}
 &  1 & 0.124\pmm0.168 & 0.304\pmm0.195 & 0.000\pmm0.340 & 0.180\pmm0.159 & \textbf{0.486\pmm0.235} \\
 &  3 & 0.147\pmm0.179 & 0.461\pmm0.227 & 0.008\pmm0.376 & 0.332\pmm0.205 & \textbf{0.790\pmm0.277} \\
 &  7 & 0.270\pmm0.179 & 0.547\pmm0.239 & 0.140\pmm0.463 & 0.312\pmm0.175 & \textbf{0.839\pmm0.289} \\
 & 10 & 0.308\pmm0.168 & 0.668\pmm0.279 & 0.153\pmm0.329 & 0.534\pmm0.227 & \textbf{0.925\pmm0.278} \\
 & 15 & 0.401\pmm0.169 & 0.721\pmm0.202 & 0.082\pmm0.301 & 0.513\pmm0.211 & \textbf{0.991\pmm0.272} \\
\midrule
{} & {\textit{Metrics}} & \multicolumn{5}{c}{\textit{Action-Matching}} \\
\midrule
\pix\texttt{MIMIC-III-2a}
 & ACC & 0.861\pmm0.013 & 0.872\pmm0.007 & 0.865\pmm0.007 & 0.875\pmm0.004 & \textbf{0.891\pmm0.004} \\
 & AUC & 0.914\pmm0.003 & 0.911\pmm0.007 & 0.906\pmm0.003 & 0.915\pmm0.001 & \textbf{0.922\pmm0.004} \\
 & APR & 0.902\pmm0.005 & 0.898\pmm0.006 & 0.885\pmm0.001 & 0.904\pmm0.002 & \textbf{0.912\pmm0.005} \\
\midrule
\pix\texttt{MIMIC-III-4a}
 & ACC & 0.696\pmm0.006 & 0.701\pmm0.007 & 0.682\pmm0.005 & 0.707\pmm0.005 & \textbf{0.720\pmm0.007} \\
 & AUC & 0.859\pmm0.003 & 0.864\pmm0.003 & 0.857\pmm0.002 & 0.864\pmm0.002 & \textbf{0.873\pmm0.002} \\
 & APR & 0.659\pmm0.007 & 0.667\pmm0.006 & 0.665\pmm0.003 & 0.673\pmm0.003 & \textbf{0.681\pmm0.008} \\
\bottomrule
\end{tabular}